%% file: arxiv.tex
\documentclass{applemlr}

\input{apple_preamble}

\title{\input{tex/title}}

\author[1]{Ozan Sener}
\author[1,*]{Anand Avati}
\author[1,*]{David Hafner}
\author[1,*]{Jierui Lin}
\author[1,*]{Laura Manduchi}
\author[1,*]{Xiaoying Pang}
\author[1,2]{Langwen Huang}
\author[2]{Torsten Hoefler}
\author[1,3]{Russ Schumacher}
\author[1]{Vladlen Koltun}

\affiliation[1]{Apple}
\affiliation[2]{ETH Zurich}
\affiliation[3]{Colorado State University}

\abstract{\input{tex/abstract}}

\metadata[Contribution]{\sffamily Authors with $^*$ contributed equally to this work.}
\metadata[Correspondence]{\sffamily Ozan Sener: \url{ozansener@apple.com}}
\date{\sffamily\today}

\input{tex/common}

\makeatletter
\let\orig@paragraph\paragraph
\renewcommand{\paragraph}{\@ifstar{\@paragraph@star}{\@paragraph@nostar}}
\newcommand{\@paragraph@star}[1]{\orig@paragraph*{#1.}}
\newcommand{\@paragraph@nostar}[1]{\orig@paragraph{#1.}}
\makeatother
\newcommand{\Parasplit}{}
\let\cite\citep

\begin{document}

\maketitle

\section*{Introduction}
\input{fig_texas}
\input{fig_missouri}

\input{tex/intro}
\input{fig_model}
\input{tex/fitzroy}

\input{fig_score_cards}
\input{fig_realism}
\input{tex/results}
\input{tex/conclusion}

\bibliographystyle{plainnat}
\bibliography{refs}

\appendix

\setcounter{figure}{0}
\setcounter{table}{0}
\setcounter{equation}{0}
\renewcommand{\thefigure}{S\arabic{figure}}
\renewcommand{\thetable}{S\arabic{table}}
\renewcommand{\theequation}{S\arabic{equation}}

\section*{Supplementary Information}
\input{tex/methods}
\input{tex/suppl_text}
\input{tex/suppl_figures}

\input{tex/suppl_tables}

\end{document}

%% file: apple_preamble.tex
\usepackage{amsmath}
\usepackage{enumerate}
\usepackage{algorithm}
\usepackage{algpseudocode}
\usepackage{amsfonts}
\usepackage{amsthm}
\usepackage{cleveref}
\usepackage{diagbox}
\usepackage{colortbl}
\usepackage{amssymb}
\usepackage{xspace}
\usepackage{wrapfig}
\usepackage{adjustbox}
\usepackage{tabularx}
\usepackage{booktabs}
\usepackage{mathtools}
\usepackage{tikz}
\usepackage{enumitem}
\usepackage{silence}
\usepackage{dsfont}
\usepackage[table]{xcolor}
\usepackage[dvipsnames]{xcolor}
\usepackage{multirow}
\usepackage{makecell}
\usepackage{xfakebold}
\input{math_commands}

\definecolor{textgray}{HTML}{6E6E73}
\usetikzlibrary{positioning, calc}
\usetikzlibrary{decorations.pathmorphing}

\makeatletter
\patchcmd{\wrong@fontshape}{\@gobbletwo}{}{}{}
\makeatother
\numberwithin{equation}{section}
\makeatletter
\AtBeginDocument{
  \urlstyle{sf}
  
}
\makeatother

\definecolor{light}{RGB}{125, 125, 125}
\crefname{tcb@cnt@pbox}{code}{code}
\Crefname{tcb@cnt@pbox}{Code}{Code}
\crefname{assumption}{assumption}{assumption}
\Crefname{assumption}{Assumption}{Assumptions}

\newtcolorbox[auto counter]{pbox}[2][]{
  colback=white,
  title=Code~\thetcbcounter: #2,
  #1,fonttitle=\sffamily,
  fontupper=\sffamily,
  arc=2pt,
  colframe=bgcolor,
  coltitle=fgcolor,
  colbacktitle=bgcolor,
  toptitle=0.25cm,
  bottomtitle=0.125cm
}

\makeatletter
\newcommand\applefootnote[1]{%
  \begingroup
  \renewcommand\thefootnote{}%
  \renewcommand\@makefntext[1]{\noindent##1}%
  \footnote{#1}%
  \addtocounter{footnote}{-1}%
  \endgroup
}
\makeatother

\definecolor{cverbbg}{gray}{0.90}

%% file: math_commands.tex
\usepackage{amsmath,amsfonts,bm}

\def\eqref#1{equation~\ref{#1}}

\def\1{\bm{1}}

\DeclareMathAlphabet{\mathsfit}{\encodingdefault}{\sfdefault}{m}{sl}
\SetMathAlphabet{\mathsfit}{bold}{\encodingdefault}{\sfdefault}{bx}{n}



%% file: tex/title.tex
Skillful and realistic precipitation forecasting with reinforced diffusion

%% file: tex/abstract.tex
Accurate precipitation forecasting remains elusive as physics-based models struggle to resolve mesoscale convection and machine learning often yields physically inconsistent, blurred results that consistently underestimate extreme events. We present FitzRoy, the first system that simultaneously achieves state-of-the-art predictive skill and physical realism over the continental United States. Through a reinforced diffusion framework, where a conditional discrete diffusion model is refined via reinforcement learning, FitzRoy effectively resolves the long-standing trade-off between pointwise accuracy and structural fidelity. Across lead times up to 24 hours, FitzRoy consistently matches or exceeds state-of-the-art models while preserving structural realism and capturing extreme events. Its low-latency inference further enables deployment in real-time operational settings. In case studies such as the 2025 Texas Hill Country floods and the 2026 Missouri floods, FitzRoy signals record-breaking accumulations that existing models largely missed. By bridging the gap between realism and skill, FitzRoy establishes a new frontier for high-fidelity precipitation forecasting.

%% file: tex/common.tex
\usepackage{booktabs}
\usepackage{makecell}%
\usepackage{multirow}%
\usepackage{nicefrac}

\makeatletter
\g@addto@macro\UrlBreaks{\do\-}
\makeatother

\newcolumntype{C}[1]{>{\centering\let\newline\\\arraybackslash\hspace{0pt}}m{#1}}
\newcommand{\ind}[1]{\mathbf{1}_{\{ #1 \}}} 
\newcommand{\smooth}[1]{\mathbf{S}_{\mathcal{N}}\!\left( #1 \right)} 
\newcommand{\leadtime}{\tau}

\DeclareRobustCommand\onedot{\futurelet\@let@token\@onedot}
\def\@onedot{\ifx\@let@token.\else.\null\fi\xspace}

%% file: fig_texas.tex
\begin{figure*}[t!]
    \centering
    \includegraphics[width=16.59cm]{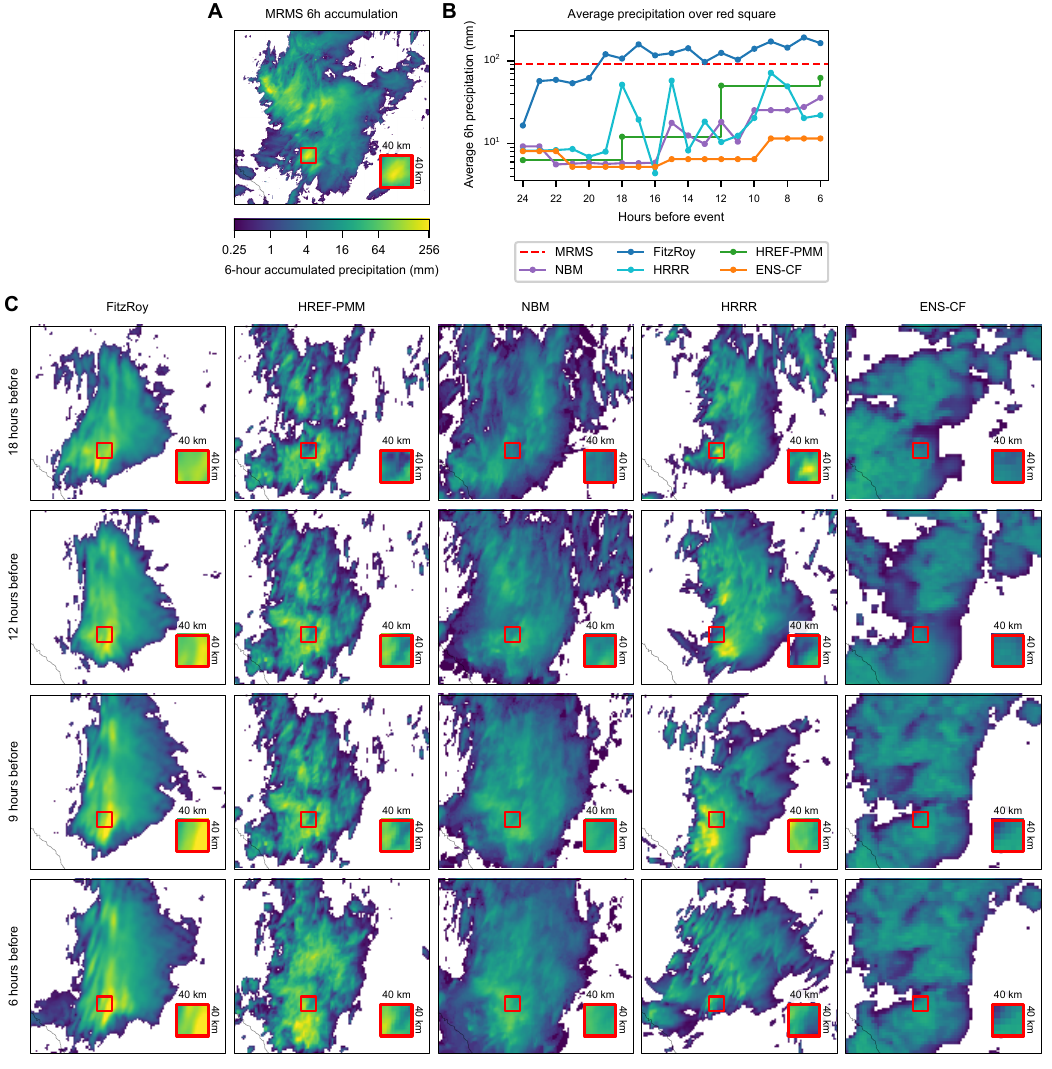}
    \caption{
        Flooding in the Texas Hill Country.
        (\textbf{A})~6-hour observed accumulated precipitation from radar between 03:00 and 09:00 UTC on 4 July 2025 around Camp Mystic (center of red box). 
        (\textbf{B})~Area-averaged observed and forecasted accumulations over the red box shown in panel A (Lat: $29.8^\circ$ -- $30.2^\circ$, Lon: $-99.6^\circ$ -- $-99.2^\circ$) for different initialization times. 
        (\textbf{C})~6-hour accumulation forecasts for different initialization times (lead times). While physics-based systems such as HRRR struggle even at 6-hour lead times, FitzRoy consistently predicts extreme precipitation 19 hours in advance. FitzRoy slightly overpredicts the rain, but for an extreme event this overprediction is preferable to underprediction for emergency alertness. FitzRoy also maintains consistency across initialization times, whereas HRRR degrades as the lead time shortens and the deadly flooding approaches.
    }
    \label{fig:texas_flood}
\end{figure*}

%% file: fig_missouri.tex
\begin{figure*}[t!]
    \centering
    \includegraphics[width=16.59cm]{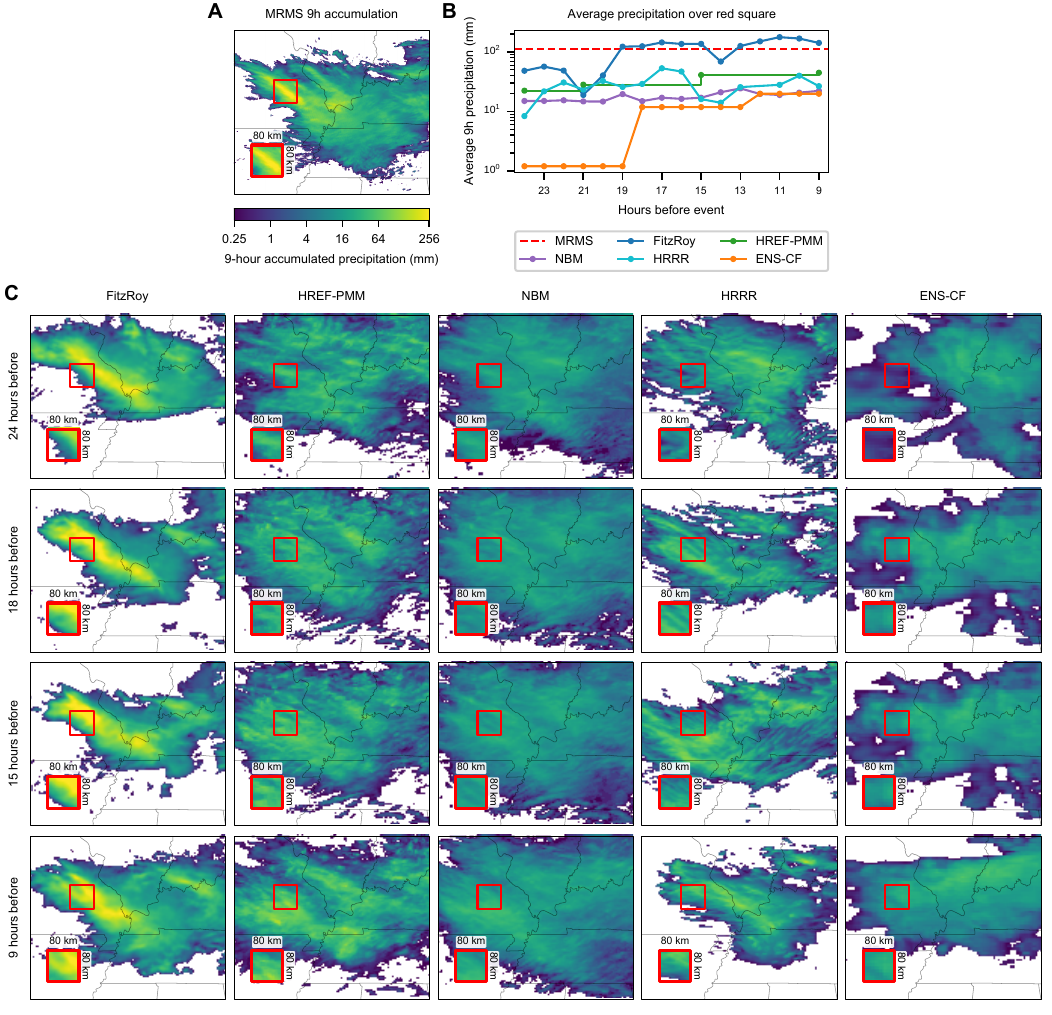}
    \caption{
    Missouri flood.
    (\textbf{A})~9-hour observed accumulated precipitation from radar between 03:00 and 12:00 UTC on 10 July 2026.  
    (\textbf{B})~Area-averaged observed and forecasted accumulations over the red box shown in panel A (Lat: $37.28^\circ$ -- $38.00^\circ$, Lon: $-91.48^\circ$ -- $-90.56^\circ$) for different initialization times. 
    (\textbf{C})~9-hour accumulation forecasts for different initialization times (lead times). FitzRoy consistently identifies the extreme accumulation magnitude across multiple initialization times, whereas operational baselines exhibit significant underprediction.}
    \label{fig:missouri_flood}
\end{figure*}

%% file: tex/intro.tex
Accurate precipitation forecasting is one of the most important problems in meteorology with direct implications for water resource management, agricultural stability, and disaster resilience \cite{Bauer_2015}. Despite decades of advancement in numerical weather prediction (NWP), short-range forecasts of high-intensity, localized rainfall often lack the precision required for effective emergency response. This stems from a fundamental physical constraint: NWP models typically operate at resolutions too coarse to explicitly resolve key processes such as deep convection, cloud microphysics, and boundary-layer turbulence. Furthermore, precipitation is not a fundamental prognostic variable of atmospheric equations; instead, it is diagnosed from thermodynamic and dynamical fields via highly idealized, heuristic parameterizations. Consequently, even state-of-the-art physics-based systems struggle to capture the sharp gradients and stochastic nature of mesoscale convective systems~\cite{HRRRpart2}.

Machine learning (ML) has recently matched and even surpassed NWP in predicting smooth variables such as temperature and geopotential height by exploiting reanalysis datasets and flexible architectures~\cite{Aurora, GraphCast, GenCast, FourCastNet, FuXi, Pangu, aadvark, FGN, neuralgcm}. However, these systems struggle to forecast precipitation and often fail to resolve the sharp, mesoscale structures necessary for predicting high-impact events like flash floods. Efforts to address this via local grid-to-grid models targeting radar observations, such as MetNet \cite{MetNet, MetNetv2, MetNetv3}, DGMR~\cite{dgmr}, and NowcastNet~\cite{nowcastnet}, show skill at very short lead times but often degrade into unrealistically smooth forecasts as lead time increases. Conversely, emulators of convection-allowing models such as StormCast and HRRRCast \cite{StormCast, hrrrcast} can generate high-fidelity patterns but typically fail to match the skill of physics-based models. Recent systems such as WoFSCast and StormScope extend ML forecasting toward storm-scale dynamics, but remain similarly limited in forecast horizon ($<12$ h)~\cite{wofscast,stormscope}, while StormScope does not directly predict precipitation accumulation~\cite{stormscope}.

A primary obstacle remains the fundamental tension between predictive skill and physical realism. While physics-based models often sacrifice skill to maintain physical consistency, standard ML models tend to ``hedge'' their 
predictions resulting in blurred, low-amplitude rain fields that lack structural fidelity and consistently underestimate extreme events~\cite{Radford_AIWP_2025, Davis_MODE_2006a,double_penalty}. 
\Parasplit
To date, the field has lacked a system capable of delivering precipitation forecasts that are simultaneously high-skill and high-fidelity~\cite{Bonavita_2024, Radford_AIWP_2025}.

Here we introduce FitzRoy\footnote{Named after Vice-Admiral Robert FitzRoy (1805--1865), a pioneer of meteorology who established the UK Met Office and coined the term ``weather forecast''~\cite{Gribbin2004}.}, a machine learning framework that overcomes the trade-off between skill and realism to provide precipitation forecasts for the continental United States (CONUS) with lead times up to 24 hours. FitzRoy is, to our knowledge, the first system to surpass operational physics-based models while maintaining high-frequency structural realism at such a long forecast horizon. This is achieved through a reinforced diffusion framework, a generative modeling recipe that combines conditional discrete diffusion with reinforcement learning (RL). We first train a diffusion model to learn the distribution of physically plausible precipitation structures, then fine-tune the system using RL to align these generations with observed outcomes. By treating the diffusion process as a stochastic policy and ranking realizations against observed events, we apply policy gradients to bias the model toward high-skill trajectories without sacrificing structural fidelity.

We evaluate FitzRoy across held-out data spanning multiple seasons, against both in-situ station observations and radar estimates. Unlike recent models~\cite{GenCast, GraphCast, Aurora} that rely on retrospective reanalysis that is unavailable operationally, FitzRoy is trained on heterogeneous data sources and evaluated using operational data streams, ensuring its utility in operational forecasting. Consequently, FitzRoy consistently matches or outperforms strong physics-based and ML baselines in forecasting precipitation probability, rate, and accumulation at lead times up to 24 hours. The system is fully operational and low-latency, generating high-resolution forecasts within seconds. To verify structural realism, we analyze power spectral density and calibration, demonstrating that FitzRoy's forecast distributions closely mirror the statistical properties of observations. Case studies of high-impact events further demonstrate tangible benefits; notably, during the catastrophic Texas Hill Country floods of July 2025~\cite{cnn2025texasflood} (Fig.~\ref{fig:texas_flood}) and Missouri floods of July 2026 (Fig.~\ref{fig:missouri_flood}), FitzRoy accurately signals the potential for extreme accumulations 18--20 hours in advance, whereas state-of-the-art models underpredict the event's magnitude and misplace its spatial footprint. Notably, the Missouri floods occurred approximately two years after the cutoff of FitzRoy's training data, providing evidence of temporal generalization to extreme events well beyond the training period.

%% file: fig_model.tex
\begin{figure*}[t!]
    \centering
    \includegraphics[width=16.59cm]{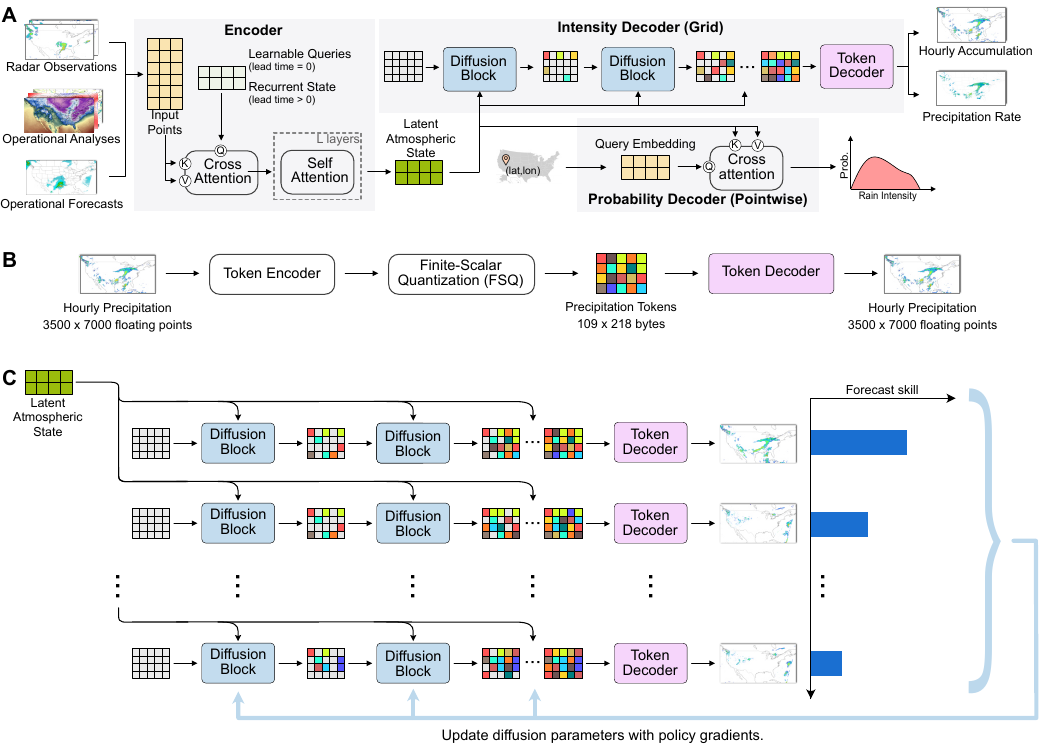}
    \caption{
        Overview of FitzRoy.
        (\textbf{A})~ An encoder with cross-attention ingests heterogeneous inputs and produces a latent atmospheric state. A pointwise probability decoder queries this state at specific locations to generate probabilistic forecasts recurrently for each lead time, while a gridwise intensity decoder maps the latent state to a spatially coherent precipitation-intensity field over the full grid via discrete diffusion.  For clarity, only a single lead time is visualized.
        (\textbf{B})~To represent high-fidelity precipitation structure faithfully, we learn a tokenizer that maps precipitation fields to a discrete vocabulary, and train the joint decoder in this token space. 
        (\textbf{C})~Diffusion objectives encourage realism but do not directly optimize forecast skill. We therefore fine-tune with reinforcement learning to improve skill while preserving realism: we sample multiple precipitation fields from the diffusion model, rank them, and apply policy gradients to bias sampling towards high-skill forecasts.
    }
    \label{fig:model}
\end{figure*}

%% file: tex/fitzroy.tex

\section*{FitzRoy}
FitzRoy is a unified machine learning framework designed to simultaneously predict (i) accumulated precipitation, (ii) instantaneous precipitation rates, and (iii) the exceedance probability of critical thresholds. As illustrated in Fig.~\ref{fig:model}A, FitzRoy first maps disparate observations into a high-dimensional latent representation of the atmosphere. This representation is intentionally decoupled from rigid grid-based assumptions to support diverse input modalities and forecasting objectives. 

To hedge against localization uncertainty, standard training objectives favor spatially smooth, ``blurred'' forecasts. While these achieve high skill scores in pointwise metrics, they often fail to resolve the mesoscale organization and sharp gradients required for accurate hydrologic modeling~\cite{Bonavita_2024}. 
To address this, FitzRoy employs two specialized decoders that take the latent atmospheric state as input: a probability decoder for point-level probabilities and a generative intensity decoder for precipitation fields. We adopt a two-stage training curriculum. In the first stage, we optimize the latent representation and the probability decoder. In the second stage, we train the intensity decoder conditioned on the latent representation via a reinforced diffusion framework that reconciles predictive skill with structural realism.

\subsection*{Learning the latent atmospheric representation}

Atmospheric data are intrinsically heterogeneous, comprising asynchronous radar and satellite retrievals, irregular in-situ station records, and gridded analysis fields. To overcome the rigid structural constraints inherent in traditional numerical weather prediction and grid-based ML models, we treat all inputs, including variables on regular grids, as unordered sets of points, each embedded with its coordinates in space and time. This allows for the direct ingestion of inputs with disparate coverage and sampling. Inputs include atmospheric variables (ECMWF Reanalysis v5 (ERA5)~\cite{era5} for training, Global Forecast System (GFS)~\cite{gfs} forecasts for inference), radar-derived precipitation from the Multi-Radar/Multi-Sensor system~(MRMS)~\cite{mrms}, and forecasts from the National Blend of Models (NBM)~\cite{NBM}, encompassing surface variables and column features across multiple pressure levels.

The encoder processes the input points through cross-attention, using learnable vectors as queries and input points as keys and values. The resulting latent states are rolled forward recursively $t$ times, where $t$ is the desired lead time: at each step, the model attends over the last two latents via cross-attention, then refines the result through deep self-attention layers. The accumulated latent state represents the evolving atmospheric context as a collection of high-dimensional vectors, untethered to any specific grid. By avoiding fixed spatial grids or Markovian temporal constraints, FitzRoy captures the multi-scale, non-linear dynamics governing precipitation.

The probability decoder applies cross-attention between the latent state and queries derived from target coordinates (latitude and longitude). Precipitation is modeled as a discrete distribution and trained using cross-entropy loss against observations, directly optimizing point-wise accuracy. This stage is trained on a multi-decade dataset spanning 1990 to mid-2024.

\subsection*{Intensity forecasts via reinforced diffusion}
Precipitation exhibits spatial structure that cannot be captured by pointwise modeling alone; e.g.\ mesoscale organization, filamentary bands, cellular convection, and quasi-stationary features. The intensity decoder represents them faithfully through three integrated components: discrete tokenization, conditional diffusion, and RL-based optimization.

\paragraph*{Discrete tokenization} 
To preserve the spatial fidelity of precipitation patterns, we compress precipitation fields into a discrete latent vocabulary using a specialized tokenizer (Fig.~\ref{fig:model}B). We employ a convolutional autoencoder with a finite scalar quantization \cite{fsq} bottleneck to train this tokenizer. This reduces the problem dimensionality while ensuring that any sequence of tokens decodes into a realistic precipitation structure. In addition to enhancing realism, tokenization provides significant engineering advantages through data compression.

\paragraph*{Conditional discrete diffusion}
Using the vocabulary learned by the tokenizer, the gridwise intensity decoder models the distribution of future precipitation fields via conditional diffusion in token space~\cite{diff_book,discrete_diffusion,maskgit}. Conditioned on the latent atmospheric state, a discrete diffusion model iteratively generates realistic, spatially coherent precipitation fields for both accumulation and instantaneous precipitation rate. During training, we adopt a masked-token objective where the model learns to predict the identity of hidden tokens by attending to both the surrounding visual context and the global atmospheric state via cross-attention. At inference, the system generates forecasts through iterative unmasking via a Halton scheduler~\cite{halton}.

\paragraph*{Reinforcement learning}
While conditional diffusion effectively captures the distribution of physically realistic weather patterns, it does not explicitly optimize for forecast skill. We bridge this gap through reinforcement learning (Fig.~\ref{fig:model}C).
By treating the diffusion model as a stochastic policy, we fine-tune the system via RL. During this phase, the model generates multiple candidate realizations for a given atmospheric state. We rank those realizations based on their alignment with observed precipitation events using a composite reward combining the fractions skill score (FSS) and frequency bias at multiple intensity thresholds (see Supplementary Information). We then apply policy gradients to bias the diffusion process toward high-skill trajectories. This ensures that the resulting forecast distribution is not only visually and statistically indistinguishable from observations, but also yields individual samples with higher precision in the timing and placement of extreme events.  While recent frameworks such as DiffusionNFT~\cite{diff_nft} and Flow-GRPO~\cite{flow_grpo} explore approaches largely tailored to continuous diffusion, the discrete nature of our model allows for a simpler Reinforce-Leave-One-Out (RLOO)~\cite{rloo} strategy. To produce a single operational forecast, we scale down the sampling temperature of the diffusion process.

%% file: fig_score_cards.tex
\begin{figure*}[t!]
    \centering
    \includegraphics[width=16.59cm]{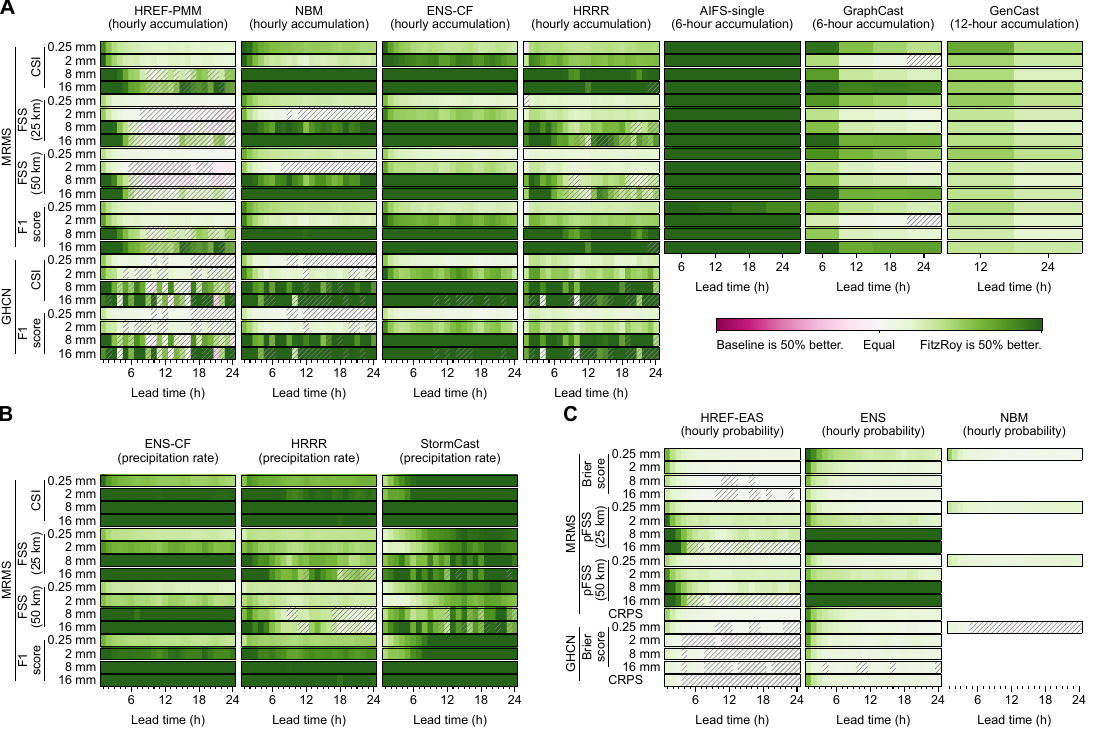}
    \caption{
        Evaluation of deterministic and probabilistic forecast skill via scorecards.
        (\textbf{A})~Precipitation accumulation. 
        (\textbf{B})~Instantaneous precipitation rate. 
        (\textbf{C})~Probability of rain.
        Each scorecard displays a collection of metrics over all forecast lead times, where hatched cells indicate statistically not significant differences (two-sided test at a 5\% significance level following~\citet{Hamill1999}). 
        Across nearly all tasks, metrics, and lead times, FitzRoy matches or outperforms the physics-based and ML baselines within their intended scope of operation.
    }
    \label{fig:results}
\end{figure*}

%% file: fig_realism.tex
\begin{figure*}[t!]
    \centering
    \includegraphics[width=16.59cm]{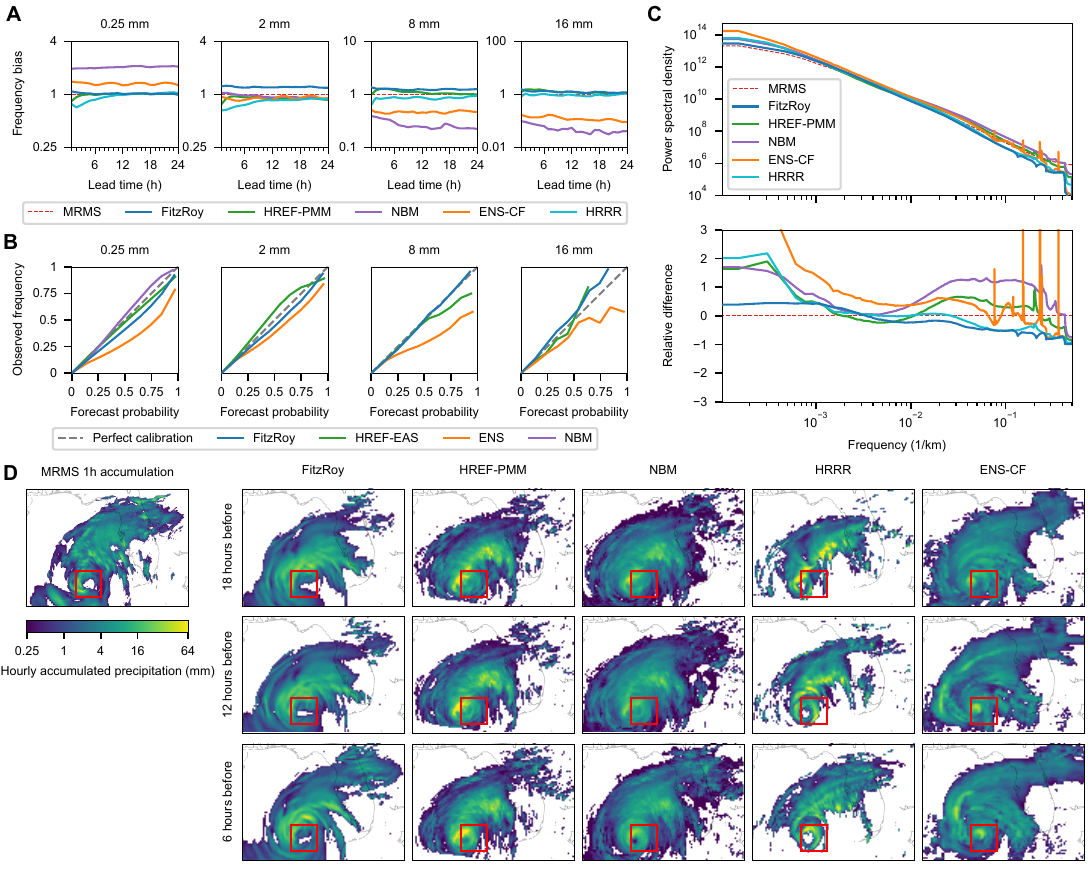}
    \caption{
        Evaluation of forecast realism.     
        (\textbf{A})~Frequency bias as a function of lead time. FitzRoy is essentially unbiased regarding rain vs. no-rain occurrence and reproduces event frequencies accurately,
        showing only slight overprediction for heavy rain. FitzRoy demonstrates superior frequency bias compared to NBM and ENS-CF, both of which significantly underpredict heavier rainfall events.
        (\textbf{B})~Calibration curves for FitzRoy and probabilistic baselines at four precipitation thresholds. The x-axis shows forecast probability and the y-axis the observed frequency of events exceeding the threshold. Perfect calibration follows the diagonal; points below (above) indicate overconfidence (underconfidence). Only bins with at least 1,000 samples are shown. FitzRoy is well-calibrated across all intensity thresholds.
        (\textbf{C})~Top:~mean power spectral density of hourly precipitation accumulation as a function of spatial frequency. Bottom:~relative difference with respect to MRMS. Curves closer to MRMS indicate better performance. FitzRoy closely matches the MRMS power spectrum, whereas all other methods overpredict power at low spatial frequencies.
        (\textbf{D})~Hurricane Milton case study (12:00 UTC October 9th, 2024). FitzRoy and HRRR both capture the hurricane structure and eye, whereas all other evaluated methods blur these features. FitzRoy localizes the storm more accurately than HRRR, which misplaces the eye of the hurricane, especially for longer lead times. 
    }
    \label{fig:results:realism}
\end{figure*}

%% file: tex/results.tex
\section*{Evaluation of forecast skill}
\label{sec2}

FitzRoy offers three main forecasting products: (i)~hourly rain probability, (ii)~instantaneous precipitation rate at the top of the hour, and (iii)~hourly precipitation accumulation. We evaluate FitzRoy against physics-based and ML baselines on a strictly held-out test period comprising forecasts from August 2024 onwards, a period never seen during training or validation. We use MRMS~\cite{mrms} and in-situ station data from GHCN~\cite{ghcn} as reference. For hourly rain probability, we compare against three ensemble systems: the High Resolution Ensemble Forecast~(HREF)~\cite{HREF}, the National Blend of Models~(NBM)~\cite{NBM} from NOAA, and the ECMWF ensemble prediction system~(ENS)~\cite{ENS}. For precipitation rate, we evaluate against the control forecasts from ENS, the NOAA High-Resolution Rapid Refresh~(HRRR) model~\cite{HRRR} and the ML-based forecasts from StormCast~\cite{StormCast}. For hourly accumulation, we use forecasts from HREF, NBM, ENS, and HRRR as benchmarks. 
In addition, we compare against the ML models AIFS-single from ECMWF~\cite{aifs}, GraphCast~\cite{GraphCast}, and GenCast~\cite{GenCast}\footnote{For this comparison, GraphCast and GenCast are initialized using non-operational ERA5 reanalysis data. This introduces information that would be unavailable in a real-time operational environment. In contrast, the evaluated FitzRoy forecasts rely exclusively on operational data. Consequently, these evaluation results should be interpreted as an upper bound on the performance of GraphCast and GenCast rather than a direct operational comparison.}. 
Since the latter three models do not provide hourly accumulations, we add up our hourly predictions to 6- and 12-hour accumulations for comparison.

Due to the varying characteristics of the above baselines (forecast frequency, accumulation window, etc.) we present pairwise scorecards in Fig.~\ref{fig:results}, where each scorecard evaluates FitzRoy within the intended scope of operation of the corresponding baseline. For each lead time, a cell in the scorecard shows the relative difference of FitzRoy compared to the corresponding baseline as well as the statistical significance following \citet{Hamill1999}. We provide metrics definitions and further evaluation details in the supplementary materials.

Across the evaluated deterministic and probabilistic precipitation tasks, FitzRoy is consistently competitive with or better than all baselines, although the magnitude of the improvement varies across models, metrics, thresholds, and lead times. The largest gains are observed relative to strong operational convection-allowing models such as HRRR and ENS-CF, as well as ML-based forecasting systems, including AIFS-single, GraphCast, and GenCast, which FitzRoy outperforms by substantial margins across most metrics. Against NBM, whose forecast is used as an input to the FitzRoy model, we similarly observe large gains across most metrics, demonstrating that FitzRoy adds substantial predictive skill beyond the forecast information it receives. Against the strong operational ensemble system HREF, the improvements are generally more modest, and some individual comparisons are statistically indistinguishable; nevertheless, FitzRoy consistently matches or exceeds HREF across the corresponding evaluation regimes. Similar behavior is observed for instantaneous precipitation rate and probabilistic forecasting, where FitzRoy remains competitive with or improves upon the corresponding operational baselines. To our knowledge, this is the first ML-based system to achieve this level of performance across both deterministic and probabilistic precipitation forecasting, effectively closing the historical performance gap between ML approaches and operational physics-based systems while exceeding existing ML baselines by considerably larger margins.

We observe the most significant performance gains in forecasting rain intensities (Figs.~\ref{fig:results}A and B). Relative to both deterministic and ensemble baselines, FitzRoy delivers superior discrimination between rainy and non-rainy events and substantially improved skill for heavy-precipitation thresholds. While operational probabilistic systems are often constrained by small ensemble sizes and the need for ad hoc dispersion inflation, FitzRoy's uncertainty emerges naturally from its probabilistic components. This design yields sharper, more informative distributions that maintain calibration even in the tails.

To evaluate the pure forecasting quality, we neglect here the data-delivery and model runtime latencies. However, in real-world operational environments, model latency is a critical bottleneck. We therefore provide a latency-aware evaluation in supplementary materials (Fig.~S1), which adjusts the effective forecast lead time according to model initialization frequency, data availability, and runtime. Because FitzRoy can ingest recent observations and produce forecasts with sub-hour latency, this operationally aligned evaluation further increases its advantage over traditional numerical weather prediction systems, whose most recent available forecasts may already be several hours old at the time of use.\footnote{We also evaluate FitzRoy against MetNet-3~\cite{MetNetv3} in the supplementary materials. Since neither source code nor operational forecasts for MetNet-3 are publicly available, we customize the evaluation protocol in an attempt to reproduce their reported results.}

To better understand the contribution of individual components to FitzRoy's performance, we present analyses in the supplementary materials. Specifically, we study the contribution of NBM forecasts, the role of reinforcement learning, and key architectural and modeling choices.

\section*{Evaluation of forecast realism}

Beyond aggregate skill scores, a precipitation model must provide faithful spatial structure to be meteorologically useful. We assess structural realism through power spectral density (PSD) analysis, calibration diagnostics, and qualitative analysis (Fig.~\ref{fig:results:realism} and supplementary materials).

FitzRoy exhibits well-calibrated probabilistic and intensity forecasts across four intensity thresholds spanning light to heavy rain~(Fig.~\ref{fig:results:realism}B). Forecast probabilities indicate no systematic over- or underconfidence. Intensity calibration, assessed via frequency bias, shows similar performance~(Fig.~\ref{fig:results:realism}A): FitzRoy is unbiased for rain/no-rain events and yields event frequencies consistent with observations for light and heavy precipitation. 

In terms of spatial realism, FitzRoy closely matches the observed MRMS spectrum across spatial scales (Fig.~\ref{fig:results:realism}C), preserving variance from convective-scale features to mesoscale organization. Despite the tendency of ML models to produce overly smooth precipitation fields, FitzRoy preserves fine-scale structures evident in radar imagery and produces power spectra that are similar to, and at some scales even closer to observations than, the physics-based HRRR and HREF systems.

These aggregate diagnostics translate into more realistic depictions of extreme weather phenomena. A qualitative analysis of 2024 Hurricane Milton~(Fig.~\ref{fig:results:realism}D) demonstrates that FitzRoy captures the storm's eye and spiral rainbands with high precision. While HRRR captures similar structures, its position errors grow at 12--18 hour lead times; conversely, HREF tends to blur these features. FitzRoy's combination of accuracy and faithful representation of the fine-scale structure demonstrates that it can match, and in some cases even surpass, the realism of convection-allowing NWP systems.

\section*{Case studies: forecasting catastrophic extremes}

To complement the quantitative analysis, we examine several high-impact events from the held-out test set and compare FitzRoy's performance in a latency-aware setting against the baselines. Our analysis focuses on: 
(i)~the flooding in the Texas Hill Country on 3--4 July 2025; 
(ii)~Missouri rainstorm on 9--10 July 2026; 
(iii)~Hurricane Milton on 9--10 October 2024; 
(iv)~flooding in Connecticut and New York on 18--19 August 2024; 
(v)~remnants of Tropical Storm Chantal on 6--7 July 2025;
(vi)~localized heavy rainfall in Ruidoso, New Mexico, on 8 July 2025; and (vii)~the northern US derecho on 20--21 June 2025.

These cases span diverse meteorological regimes, including tropical cyclones, organized convective systems, and quasi-stationary convection. In the main text, we highlight the Texas Hill Country event~(Fig.~\ref{fig:texas_flood}) and flooding in Missouri~(Fig.~\ref{fig:missouri_flood}), both of which resulted in historic flooding. We evaluate Hurricane Milton (Fig.~\ref{fig:results:realism}) as part of our realism evaluation and provide the remaining case studies in the supplementary materials.

Fig.~\ref{fig:texas_flood} compares FitzRoy forecasts with several operational NWP models for July 3--4 2025, when deadly and destructive flooding occurred in the Texas Hill Country. During this event, clusters of convective cells repeatedly formed near a mesoscale vortex and moved slowly over the same region, producing rainfall accumulations exceeding 300~mm in the early morning hours of July 4th~\cite{galarneau_TXflood}. Public officials later cited limited information about the potential for such extreme precipitation as a contributing factor to challenges in warning and emergency response~\cite{cnn2025texasflood}.

For forecasts initialized at 18:00~UTC July 3rd, approximately 12 hours before the onset of the heaviest rain, the convection-allowing HRRR (Fig.~\ref{fig:texas_flood}C, fourth column) substantially underpredicts the accumulated precipitation and fails to capture the full extent of the flood-producing swath. The HREF ensemble (Fig.~\ref{fig:texas_flood}C, second column) indicates the possibility of localized heavy rainfall in central Texas, but its maxima are displaced and its spatial footprint too limited. In contrast, FitzRoy (Fig.~\ref{fig:texas_flood}C, first column) accurately captures both the location and magnitude of the extreme accumulations, producing a coherent band of heavy precipitation aligned with the observed training convection and flood-producing region.

Crucially, FitzRoy exhibits high run-to-run consistency. While the specific location of the heaviest rain shifts between initialization times in HREF and HRRR, FitzRoy's signal remains anchored over the impacted region across multiple forecast cycles. This stability, combined with its ability to resolve mesoscale organization, illustrates how reinforced diffusion can provide actionable intelligence for high-impact events where traditional models struggle.

For another high-impact flood in Missouri on 10 July 2026~\cite{missouri2026ap}, similar themes emerge (Fig.~\ref{fig:missouri_flood}). A nearly stationary band of convection formed and produced extreme rainfall (over 200 mm in less than 12 h) and flooding in southeast Missouri. FitzRoy predicted a narrow band of heavy rainfall with intensity and structure similar to the observed storms, though with some errors in location. In contrast, other operational models underpredicted the rainfall, showed the band of convection in the wrong place, or both (Fig.~\ref{fig:missouri_flood}). As with the Texas Hill Country forecasts, FitzRoy remains consistent with a forecast of extreme precipitation in Missouri while other models predicted smaller accumulations with more run-to-run variation.

%% file: tex/conclusion.tex
\section*{Discussion}
Precipitation is among the most difficult weather variables to predict due to its extreme spatio-temporal intermittency. Historically, a ``skill-realism'' trade-off has forced a choice between the structural fidelity of high-resolution physics-based models and the statistical accuracy of regression-based machine learning models. Our results show that this trade-off can be substantially reduced. FitzRoy achieves strong performance across 24-hour lead times while preserving the high-frequency spatial structure characteristic of observed precipitation.

FitzRoy bridges this gap through reinforced diffusion. A precipitation tokenizer and conditional discrete diffusion anchor forecasts in physically plausible structures rather than the smoothed averages typical of regression-based ML models. Reinforcement learning biases the diffusion process toward high-skill trajectories, mitigating the ``double penalty'' problem, in which a realistic but slightly displaced storm is penalized more heavily than a blurred forecast. This allows FitzRoy to preserve the sharp gradients and high-frequency power spectra characteristic of real-world precipitation while achieving high accuracy in deterministic and probabilistic settings.

A key strength of our framework is its flexibility across heterogeneous inputs. Using initializations constructed from physics-based model states and observations, FitzRoy already achieves strong performance. When an additional external forecast is available, its shared latent representation provides an effective mechanism for integrating this additional information, further improving forecast quality and allowing FitzRoy to function simultaneously as a forecasting model and a learned forecast-integration system. This flexibility makes FitzRoy difficult to place within a single conventional category such as autonomous forecasting, nowcasting, downscaling, or post-processing. Instead, FitzRoy spans these regimes, with its role determined by the information available at inference time. 

Case studies such as the 2025 Texas Hill Country floods and the 2026 Missouri floods underscore the real-world utility of our approach. Operational physics-based models, constrained by their reliance on idealized parameterizations of convection, failed to signal the magnitude of the event in time for effective emergency response. FitzRoy's ability to resolve extreme convection with high run-to-run consistency represents a significant advance in forecasting high-impact, localized extremes.

Furthermore, the computational efficiency of ML models addresses a critical bottleneck: the total time to forecast. FitzRoy's low-latency inference provides actionable, high-fidelity forecasts in seconds, contrasting with the hours required to run a high-resolution physics-based ensemble. This effectively extends the actionable lead time for emergency management beyond what is currently possible with NWP alone.

Ultimately, FitzRoy demonstrates that the synthesis of skill and realism is achievable with machine learning and leads to significant operational benefits. Reinforced diffusion and shared latent representation provide a blueprint for the next generation of operational meteorology. By respecting both the underlying data distribution and the fundamental dynamics of the atmosphere, we can move toward a future where weather events are anticipated with the precision and fidelity required for effective resource management and disaster resilience.

%% file: tex/methods.tex
 \subsection*{Model details}
 FitzRoy consists of three main components: (i)~an \emph{encoder} that maps heterogeneous observations into a compact latent representation of the atmospheric state, (ii)~a \emph{probability decoder} that produces pointwise probabilistic precipitation forecasts at arbitrary query locations, and (iii)~an \emph{intensity decoder} that generates precipitation forecasts on a 1-km grid. The probability decoder is used to train the latent atmospheric representation, which subsequently serves as the conditioning signal for the intensity decoder. Training of the intensity decoder involves a tokenizer, a conditional diffusion model, and a reinforcement-learning fine-tuning stage. Each component is described in detail below.

\subsubsection*{Inputs}
We use three primary data sources as training input: ERA5 reanalysis data (21 selected atmospheric variables)~\cite{era5}, MRMS-based precipitation maps~\cite{mrms}, and precipitation forecasts from the National Blend of Models (NBM)~\cite{NBM}; see Tables~\ref{tab:train_inputs},~\ref{tab:era5-channels}, and~\ref{tab:non-era5-channels} for further details.

\subsubsection*{Targets}
Training targets include ERA5 analyses and MRMS observations. Similar to the inputs, we use the hourly accumulated precipitation from MRMS observations. Further details can be found in Tables~\ref{tab:train_targets},~\ref{tab:era5-channels}, and~\ref{tab:non-era5-channels}.

\subsubsection*{Encoder}
Given an initialization time $t_0$ and target lead time $\leadtime$, input embeddings are constructed from ERA5, MRMS, and NBM forecasts. ERA5 and MRMS are used at $t_0$ and the previous hour $t_{-1}=t_0-1$, while NBM forecasts are used through lead time $\leadtime$. During training, we apply a lead-time curriculum, sampling earlier lead times exponentially more often initially ($1000:1$ ratio) before transitioning to near-uniform sampling ($16:1$ ratio) over $20$ million samples. In operational deployments, ERA5 is replaced by Global Forecast System (GFS) forecasts, with appropriate lead times selected from the 6-hourly GFS cycles to minimize latency.

For each data source and location, the encoder constructs an embedding comprising weather variables, their Fourier features, spatial and temporal positional embeddings, and a one-hot source indicator. Spatial embeddings use Fourier features of latitude and longitude normalized to $[-1,1]$, with $128$ log-spaced frequencies (1--3500 for latitude, 1--7000 for longitude), yielding $512$ dimensions per point. Temporal embeddings separately encode hour-of-day, day-of-year, and year and are projected to $512$ dimensions. A small Gaussian positional jitter is applied to reduce dependence on absolute coordinates while preserving relative geometry.

The resulting embeddings form an unordered, variable-size set of input points. FitzRoy maps this set to a latent representation using cross-attention with learned latent queries. MRMS and ERA5 observations at $t_{-1}$ and $t_0$ are encoded into $\mathbf{z}_{-1}, \mathbf{z}_0 \in \mathbb{R}^{N \times D}$ using $N=1024$ trainable queries of dimension $D=1024$. Keys and values are linear projections of the input embeddings, each of dimension $C=1758$. This compression to a fixed-size latent state enables scalable downstream forecasting.

The encoder then evolves the latent representation recurrently over lead time. For each hour $i \in \{1,\ldots,\leadtime\}$, the latest latent serves as the query in a cross-attention block, while the two most recent latents, augmented with learned temporal-position embeddings, provide keys and values. The output is processed by $L=8$ self-attention layers to produce $\mathbf{z}_i$.

When NBM forecasts are available for hour $i$, they are independently mapped to a conditioning latent $\mathbf{c}_i$ using the same cross-attention mechanism with shared weights. The recurrent and NBM latents are combined as $\tilde{\mathbf{z}}_i=(\mathbf{z}_i+\mathbf{c}_i)/2$ entirely in learned latent space, rather than by averaging precipitation forecasts. The resulting $\tilde{\mathbf{z}}_i$ represents lead hour $i$ and is propagated to the next recurrent step, allowing NBM guidance to influence the trajectory over time. Gradients are not propagated through the NBM branch, treating NBM as external forecast guidance.

Because of sparsity, only pixels with non-zero precipitation are included. This sparsification makes cross-attention tractable while retaining information. The encoder is trained jointly with the probability decoder.

\subsubsection*{Probability decoder}
FitzRoy's probability decoder maps each latent atmospheric representation to weather variables at arbitrary spatial query locations within CONUS. For target lead time $\leadtime$, the decoder receives the corresponding latent $\tilde{\mathbf{z}}_\leadtime \in \mathbb{R}^{N \times D}$ from the encoder. Spatial queries are latitude and longitude coordinates encoded as $512$-dimensional random Fourier features of the normalized coordinates. Each query is projected to $D=1024$ and used in a cross-attention block, with the atmospheric latent providing keys and values. The output, without a residual connection to the query, passes through a feed-forward network, layer normalization, GELU activation, and a final linear projection to produce logits.

The decoder predicts variables corresponding to ERA5 analyses and MRMS observations. The primary target is hourly precipitation accumulation (mm), formulated as a classification problem. Precipitation is discretized into 400 bins of width 0.25 mm spanning $[0,100)$ mm, plus an overflow bin for values $\ge 100$ mm, for 401 bins in total. The decoder outputs logits over these bins and is trained with cross-entropy loss where observations are available. Additional channels correspond to non-precipitation weather variables, supervised with mean squared error against ERA5 analyses, and precipitation type, supervised with cross-entropy over predefined classes. The objective is a weighted sum of precipitation amount and type cross-entropy losses and grid-based regression loss.

\subsubsection*{Intensity decoder} 
The intensity decoder produces spatially coherent precipitation fields that are both realistic and accurate. We construct the intensity decoder in three main steps: (i)~learning a discrete vocabulary of precipitation fields via tokenization, (ii)~learning a conditional diffusion process in the token space, and (iii)~applying post-diffusion reinforcement learning (RL) to steer sampling toward high-likelihood, high-skill realizations.

\subsubsection*{Learning a vocabulary of realistic precipitation}
We learn a discrete vocabulary whose codes decode to plausible precipitation fields while covering realistic precipitation patterns. Mapping high-dimensional observations into this space reduces forecasting to token prediction and provides substantial compression, lowering memory and I/O costs.

FitzRoy uses a convolutional autoencoder with a quantization bottleneck as its tokenizer. The encoder applies hierarchical downsampling with ResNeXt~\cite{resnext} blocks, group normalization, and Swish activations~\cite{swish}, with self-attention at the deepest level; the decoder mirrors this architecture. Finite Scalar Quantization (FSQ)~\cite{fsq} discretizes the latent representation by rounding bounded values to predefined levels.

MRMS precipitation fields are first downsampled \(4\times\) to \(875\times1750\), followed by three \(2\times\) encoder downsampling stages producing \(437\times875\), \(218\times437\), and \(109\times218\) grids, for \(32\times\) total compression. Each location is quantized into a 7-dimensional FSQ code with 4 levels per dimension, yielding a vocabulary of 16,384 (\(4^7\)) represented by 14 bits. Padding and cropping during encoding and decoding preserve exact spatial alignment.

The loss combines (i)~binary cross entropy (BCE), (ii)~masked mean absolute error (MAE) on log-transformed intensity, and (iii)~LPIPS~\cite{LPIPS}. 

\subsubsection*{Conditional diffusion in the token space} 
Using a fixed tokenizer, we model precipitation forecasting as conditional generation in a discrete latent space following MaskGIT~\cite{maskgit} formulation. A pre-trained, frozen encoder provides the atmospheric conditioning state, while the diffusion model predicts the corresponding future token grid.

Training uses masked-token denoising: a random fraction of tokens is masked and reconstructed by a bidirectional Transformer using self-attention over the token grid and cross-attention to the atmospheric state. Each lead hour is decoded independently using the same curriculum as the probability decoder. The 480M-parameter model has 24 layers with 16-head self-attention, cross-attention to the $1024\times1024$ atmospheric latent, and GELU feed-forward layers. To reduce sequence length, the \(109\times218\) grid is padded and downsampled with a \(2\times2\), stride-2 projection to \(55\times109\), then restored with PixelShuffle and cropping. Atmospheric conditioning is dropped 10\% of the time to enable Classifier-Free Guidance (CFG), and training uses Exponential Moving Average (EMA).

At inference, generation begins from a fully masked grid and iteratively unmasks tokens according to the arccosine schedule, $\gamma(k/K)=\frac{2}{\pi}\arccos(k/K)$, where $\gamma(k/K)$ denotes the target fraction of tokens remaining masked after decoding step $k$ of $K$ total steps. A Halton scheduler~\cite{halton} selects token positions to distribute
updates spatially across the grid.

\subsubsection*{Reinforcement learning} 
While the conditional discrete diffusion model approximates the distribution of future weather, we aim to steer the sampling process toward realizations that are both physically plausible and maximize forecast skill. We treat the iterative decoding process as a sequential decision-making policy. By applying Reinforcement Learning~(RL), we reshape the sampling trajectory to concentrate probability mass on realizations that maximize a skill-based reward function.
We employ a composite reward that balances spatial skill with statistical consistency by combining the fractions skill score~(FSS), calculated over a 25-km neighborhood, with frequency bias~(FB) at four precipitation-intensity thresholds (0.25, 2, 8, and 16~mm). The component and threshold weights are provided in Table~\ref{tab:hyperparams}.
We use the batch-normalized advantage with REINFORCE leave-one-out~(RLOO)~\cite{rloo} to update the policy toward higher reward. To remain near the base diffusion policy, we add an in-policy regularizer comprising (i)~a sampled reverse-KL penalty from the current policy to the
frozen base policy, using the estimator adopted in
GRPO~\cite{grpo}, and (ii)~a cross-entropy term to the original ground-truth tokens. The complete minimized objective comprises the policy-gradient loss and these two regularization terms. Gradients are accumulated over all decoding steps and trajectories per conditioning input, which stabilizes training and optimizes the end-to-end sampling process. The reward weights, coefficients of the three loss terms, number of trajectories, and number of decoding steps are provided in Table~\ref{tab:hyperparams}.

\subsection*{Evaluation details}

We choose a total of 8 weeks (representing 4 seasons) as our hold-out evaluation set that is never seen during training or validation: 2--15 October 2024, 2--15 January 2025, 5--18 April 2025, and 2--15 July 2025. Our models are trained exclusively on data available through June 2024, ensuring that the evaluation assesses true out-of-sample generalization to future conditions.
We use the hourly multi-sensor-based MRMS rain accumulation (``MultiSensor\_QPE\_01H\_Pass2'')~\cite{mrms} and the in-situ station observations from GHCN~\cite{ghcn} as reference for our evaluations. We regrid the outputs of all models to the 1-km MRMS grid via budget interpolation, to evaluate all forecasts at the scale of the observed precipitation. Moreover, we focus our evaluation on the land region of CONUS.
Following \citet{Hamill1999}, we evaluate the statistical significance of performance differences using two-sided tests at a 5\% significance level.

\subsubsection*{Baselines}

We present a summary of the output characteristics of the considered baseline models in Table~\ref{tab:baselines}.
While the historical forecasts of some baselines can be accessed via publicly available data archives (NBM~\cite{NBM}, ENS-CF~\cite{ENS}, HRRR~\cite{HRRR}, and AIFS-single~\cite{aifs}), others require special processing. 
In particular, we construct the hourly probabilities of rain for ENS by downloading the full ENS ensemble (51 members) and computing the fraction of forecasts exceeding a particular threshold.
The HREF~\cite{HREF} ensemble outputs are not archived, only the individual ensemble members. Following technical reports~\cite{oper_models_ency,href_system} we combine them as follows: We compute the probability matched mean (PMM)~\cite{pmm} to construct an intensity map for deterministic evaluation. To produce ensemble probabilities, we implement the ensemble agreement scale~(EAS)~\cite{eas} method. We also evaluated a simple ensemble mean and the neighborhood maximum ensemble probability~(NMEP)~\cite{Roberts2019,nmep} approach, but PMM and EAS provided better results.\footnote{For the Missouri flood case study (Fig.~\ref{fig:missouri_flood}), the archive data were missing while the operational ensemble output was available and therefore used for comparison.} We generate forecasts for the ML models GraphCast~\cite{GraphCast}, GenCast~\cite{GenCast}, and StormCast~\cite{StormCast} with the code and model weights provided by the authors. We used ERA5 reanalysis data as input to GraphCast and GenCast, which are produced retrospectively and are unavailable in an operational setting. This introduces access to information beyond what would be available in an operational forecast, and thus the reported results should be viewed as indicative rather than directly comparable to strictly operational models. In contrast, StormCast uses operational data from the Global Forecast System (GFS)~\cite{gfs} as input. However, since GFS runs only every 6 hours, we input GFS forecasts for a fair evaluation in a real-time operational setting (cf.\ Fig.~\ref{fig:operational_setting} and Fig.~\ref{fig:stormcast_prate_oper}); similar to the inference mode of FitzRoy.

\subsubsection*{Critical success index (CSI)}

Given true positives (TP), false positives (FP), and false negatives (FN), the CSI is a binary categorical score:
\begin{equation}
    \text{CSI} = \frac{\text{TP}}{\text{TP} + \text{FP} + \text{FN}} \,.
\end{equation}
We evaluate CSI for four rain intensity thresholds: 0.25~mm (light rain), 2~mm (medium rain), 8~mm (heavy rain), 16~mm (very heavy rain).

\subsubsection*{Fractions skill score (FSS)}

FSS~\cite{fss} aims to counteract the double-penalty issue. To this end, it takes into account a neighborhood $\mathcal{N}$ to evaluate binary categorical decisions. Let $\ind{x_i > \lambda}$ denote the indicator function, equal to 1 when $x_i > \lambda$ and 0 otherwise. With
\begin{equation}
    \smooth{\ind{x_i > \lambda}} = \frac{1}{|\mathcal{N}_i|} \sum_{j \in \mathcal{N}_i} \ind{x_j > \lambda}
\end{equation}
the fractions skill score is then given by
\begin{equation}
    \label{eq:fss}
    \text{FSS}_{\mathcal{N},\lambda} = 1 - 
    \frac{\sum_i^N \left( \smooth{\ind{x_i > \lambda}} - \smooth{\ind{y_i > \lambda}} \right)^2}
    {\sum_i^N \left( \smooth{\ind{x_i > \lambda}}^2 + \smooth{\ind{y_i > \lambda}}^2 \right)} \,,
\end{equation}
where $N$ denotes the number of forecasts, and $x_i$ the predicted and $y_i$ the observed intensity value.
We consider the same thresholds $\lambda$ as for CSI, and focus our evaluations on square neighborhoods of size 25 and 50 km. 

\subsubsection*{F1-score}

The F1-score is the harmonic mean of precision and recall:
\begin{equation}
    \text{F1-score} 
    = \frac{2 \cdot \text{precision} \cdot \text{recall}}{\text{precision} + \text{recall}}
    = \frac{2 \cdot \text{TP}}{2 \cdot \text{TP} + \text{FP} + \text{FN}} \,.
\end{equation}

\subsubsection*{Brier score}
The Brier score is the mean squared difference between the predicted probability $P_i(x > \lambda)$ of precipitation exceeding $\lambda$ and the observation: $\text{BS}_\lambda = \frac{1}{N} \sum_i^N \left( P_i(x > \lambda) - \ind{y_i > \lambda} \right)^2 \,.$

\subsubsection*{Probabilistic fractions skill score (pFSS)}

pFSS~\cite{pFSS} is an adaptation of the FSS for probabilistic forecasts. It replaces the binary probabilities $\ind{x_i > \lambda} \in \{0, 1\}$ in Equation~\ref{eq:fss} with the tail probabilities $P_i(x_i > \lambda)$:
\begin{equation}
    \label{eq:pfss}
    \text{pFSS}_{\mathcal{N}, \lambda} = 1 - 
    \frac{\sum_i^N \left( \smooth{P_i(x_i > \lambda)} - \smooth{\ind{y_i > \lambda}} \right)^2}
    {\sum_i^N \left( \smooth{P_i(x_i > \lambda)}^2 + \smooth{\ind{y_i > \lambda}}^2 \right)} \,.
\end{equation}

\subsubsection*{Continuous ranked probability score}

The continuous ranked probability score (CRPS) is a generalization of the Brier score to continuous variables. It compares the full predicted cumulative distribution function (CDF), i.e.\ $P_i(x_i \leq z)$, against the observed outcome $y_i$:
\begin{equation}
    \text{CRPS} = \frac{1}{N} \sum_i^N \int_{0}^{\infty} \left( P_i(x_i \leq z) - \ind{ y_i \leq z } \right)^2 dz \,.
\end{equation}
Practically, we are given discretized CDFs such that CRPS is evaluated as
\begin{equation}
    \frac{1}{N} \sum_i^N \sum_{k}^{K} \left( P_i(x_i \leq z_k) - \ind{y_i \leq z_k} \right)^2 \cdot \Delta_k \,,
\end{equation}
where $\Delta_k$ denotes the width of the intensity bin and $z_k$ the associated intensity. We use $10$ logarithmically scaled intensity bins for evaluation with the following bin boundaries:
$0$, $0.25$, $0.5$, $1$, $2$, $4$, $8$, $16$, $32$, $64$, $128$ mm.

\subsubsection*{Calibration curve}

To compute the calibration curve for a certain event, e.g.\ precipitation exceeding an intensity threshold, we group all forecast probabilities into $K_p = 10$ equally spaced bins. Then, we plot the mean of the predicted probabilities within each bin on the x-axis, and the y-axis shows the observed frequency within the bins. Perfect calibration is given by the identity line; points below (above) indicate overconfidence (underconfidence). We only display bins with at least 1,000 samples.

\subsubsection*{Frequency bias}

The frequency bias for a certain rain category is given by the total number of forecasts within this category divided by the total number of observations. Hence, a perfect frequency bias is $1$.

\subsubsection*{Power spectral density}

For each intensity map, we compute the radially averaged power spectral densities. We average them over time ranges and lead times to produce the final power spectral density plot.

%% file: tex/suppl_text.tex
\subsection*{Analysis of model components}

\subsubsection*{Effects of reinforcement learning}
We evaluate the contribution of reinforcement learning (RL) by comparing the full FitzRoy model against the corresponding model before the RL stage. As shown in Fig.~\ref{fig:ablations}A, RL provides the largest performance improvement among the components considered in our ablation study. The gains are substantial for CSI and FSS, and persist throughout the 24-hour forecast horizon. Improvements are particularly pronounced for heavier precipitation, indicating that RL primarily enhances the model's ability to predict the location and intensity of less frequent, high-impact precipitation events. 

Because the RL objective includes FSS measures evaluated at precipitation thresholds used in the evaluation, we additionally test whether these improvements are restricted to the thresholds and area sizes explicitly used during optimization. We evaluate FitzRoy vs.\ a model without RL at previously unseen thresholds using CSI, F1-score, and FSS, where we also vary the area sizes for FSS (Fig.~\ref{fig:ablations}B). The improvements generalize across these unseen thresholds and metrics. In several cases, particularly for heavier precipitation, the relative gains are even larger than those observed at the reward thresholds. This indicates that RL does not merely optimize performance at the discrete decision boundaries included in its reward, but instead improves the underlying precipitation forecasts in a manner that generalizes across precipitation intensities and spatial scales.

We separately evaluate whether these improvements in forecast skill come at the expense of forecast realism in Fig.~\ref{fig:rl_realism}. The corresponding calibration, frequency-bias, power-spectral-density, and qualitative analyses show that the gain in predictive skill after RL is not accompanied by a degradation in realism.

\subsubsection*{Contribution of NBM forecasts}
FitzRoy incorporates NBM precipitation forecasts as external guidance throughout the recurrent forecast trajectory. To quantify how much of FitzRoy's performance is attributable to this guidance, we train a second model using the same architecture and training procedure but with NBM conditioning completely removed. Fig.~\ref{fig:ablations}A shows that NBM provides the second-largest contribution among the components considered in our ablation study after RL, with particularly pronounced improvements for heavier precipitation thresholds. Thus, external precipitation guidance provides substantial complementary information to the forecast learned from the atmospheric state and recent observations.

Importantly, however, FitzRoy remains a strong forecasting model in the complete absence of NBM precipitation guidance. We compare the without-NBM variant directly against HREF-PMM, NBM, ENS-CF, and HRRR in Fig.~\ref{fig:ablations}C. FitzRoy without NBM outperforms ENS-CF across all evaluated metrics and matches or outperforms HRRR across the corresponding comparisons. It also outperforms NBM itself for nearly all evaluated metrics, with the exception of CSI and FSS at 2-mm.

These experiments distinguish the forecasting capability of FitzRoy from its ability to integrate external forecast guidance. The without-NBM configuration evaluates FitzRoy as a standalone forecasting system and demonstrates that its performance does not arise simply from inheriting skill from NBM. The full FitzRoy configuration additionally exploits NBM guidance and achieves further improvements, particularly for heavy precipitation. In this sense, the two configurations evaluate complementary operating regimes of the framework: FitzRoy without NBM provides the most direct comparison to standalone forecasting systems such as ENS-CF and HRRR, whereas the full model evaluates FitzRoy's ability to combine its learned forecast with information from an existing multi-model forecast product such as NBM as a direct comparison to ensemble models like HREF.

We additionally evaluate the sensitivity of FitzRoy to the latent NBM integration coefficient. Replacing the default coefficient of $0.5$ with $0.25$ or $0.75$ has no systematic effect on forecast performance (Fig.~\ref{fig:ablations}A), indicating that the model is robust to the precise choice of this coefficient.

\subsubsection*{Discrete vs. continuous diffusion}
To determine whether the benefits of reinforced diffusion depend on discrete token generation, we replace the discrete diffusion based intensity decoder with a comparably sized continuous latent diffusion model based on the Flux architecture~\cite{flux}. This model variant directly predicts the normalized seven-channel tokenizer latent on the $109 \times 218$ grid, conditioned on the same frozen atmospheric representation. It contains 12 dual-stream and 26 single-stream Transformer layers, with six attention heads of dimension 128, for a total of approximately 489M parameters. We use $2 \times 2$ patchification and train the model using a velocity-based flow-matching objective.

For RL fine-tuning, we follow the same selective adaptation strategy, updating the normalization parameters throughout the model, the conditioning QKV projections, and selected embedding and output parameters. The continuous model performs competitively (Fig.~\ref{fig:ablations}A). Nevertheless, the final continuous model remains slightly below discrete-diffusion with RL, with the difference most apparent at higher rain thresholds. Hence, discrete-diffusion brings minor but important accuracy improvements.

\subsubsection*{Categorical vs. regression loss}
We further investigate the loss used to train the encoder. In place of the 401-bin categorical loss, we use a binary cross-entropy loss for the rain/dry classification together with a Huber loss on log-scale rain intensity, tuning the weighting between the two terms so that their magnitudes are comparable. As shown in Fig.~\ref{fig:ablations}A, conditioning the sampling model on a latent trained with the regression loss performs comparably to the categorical-loss latent. The categorical loss, however, yields probabilities of rain exceeding arbitrary thresholds, which is of considerable practical value in precipitation forecasting.

\subsubsection*{Point-based vs. grid-based encoder}
FitzRoy utilizes a point-based encoder that processes all heterogeneous meteorological inputs as an unordered set of continuous point tokens, relying on cross-attention to construct a flexible, grid-free latent representation of the evolving atmosphere. To isolate the effect of the point-based encoder, we compare FitzRoy against a model variant that replaces solely the per-step input encoder with a fixed, grid-based convolutional patch embedding. While FitzRoy attends over a variable-length, irregular set of explicitly positioned tokens, the ablation rasterizes each input into a dense grid, crops it, and applies a strided convolution to emit exactly $1024$ implicitly positioned tokens per source. Because every component downstream of the encoder remains identical, and the parameter counts remain comparable, any difference in forecast metrics is directly attributable to the input-encoding style. 

As shown in the evaluation scorecards, replacing the flexible point-cloud attention with a quantized, regular raster leads to a minor degradation in forecast skill across hourly accumulation metrics, demonstrating that explicitly supplying continuous positional data and allowing the model to attend over point sets provides a more accurate atmospheric representation. Additionally, a point-based encoder avoids artificial grid constraints to allow for the integration of irregular sources, such as in-situ station data, and provides better handling of missing observations; together with the empirical ablation results, these structural advantages justify its choice.

\subsection*{Additional hourly accumulation case studies}
To further demonstrate the combination of skill and realism in FitzRoy forecasts, we present several additional case study examples, which feature multiple spatial and temporal scales of heavy precipitation, and different types of weather systems. 


On 18--19 August 2024, record-breaking rainfall was produced by convective systems in the northeastern US, specifically in Connecticut and Long Island, NY~\cite{climategovnortheast}. Forecast models, including FitzRoy, have difficulty predicting the details of these storms. For the rainfall that caused flooding on Long Island early on 19 August, FitzRoy offers indications of large accumulations 9--12 h in advance of the event, while HREF and HRRR show less-defined bands of precipitation and smaller accumulations in this area (Fig.~\ref{fig:case-study-LI}).

During the remnants of Tropical Storm Chantal on 6--7 July 2025 (Fig.~\ref{fig:case-study-chantal}), a band of heavy precipitation occurred in North Carolina, which caused historic flooding in Chapel Hill, Durham, and surrounding areas \cite{carolinaobserverchantal}. FitzRoy predicts the location and approximate intensity of the heavy precipitation with 24 hours of lead time, which is more than 6 hours earlier than other forecast models. HRRR, for example, 24 hours in advance showed the remnants of Chantal over southern North Carolina, but FitzRoy predicts it in approximately the correct location. As with the Texas Hill Country forecasts, FitzRoy remains consistent with a forecast of extreme precipitation in central North Carolina while other models predicted smaller accumulations with more run-to-run variation. 

A cluster of convective storms led to very localized heavy rainfall and flash flooding in Ruidoso, New Mexico on 8 July 2025, resulting in three fatalities \cite{ruidosoflood}. Specific predictions of such localized storms are challenging, but FitzRoy consistently pointed to the potential for heavy precipitation near Ruidoso, and closely matched the accumulated precipitation in this area while other forecast models predicted less rainfall (Fig.~\ref{fig:case-study-ruidoso}). 

\subsection*{Precipitation rate case studies}

Precipitation rate is directly related to radar reflectivity, which is often used to visualize convective systems like squall lines and bow echoes. On 20--21 June 2025, intense convective storms moved across the northern US, which resulted in an outbreak of tornadoes and a derecho (swath of destructive winds). Fig.~\ref{fig:case-study-derecho} shows FitzRoy's forecasts of this squall line (right column), with the predicted spatial structure matching many of the features observed by radar as shown in MRMS (left column). 

FitzRoy also realistically depicts complex storm structures in the 9--10 July 2026 Missouri rainstorm, showing repeated development of intense convection in southeast Missouri (Fig.~\ref{fig:case-study-missouri-prate}). Predicted storms had high rain rates and exhibited ``back-building'' over Missouri, similar to the observed convective system.

\subsection*{Training cost}

Table~\ref{tab:training_cost} summarizes the computational cost of the four stages of the FitzRoy training pipeline. We report both wall-clock time and total GPU-hours to make the relative cost of the individual stages explicit. We report equivalent passes rather than literal training epochs since initialization times are sampled with replacement and paired dynamically with forecast lead times during training.

\subsection*{Attempted comparison with MetNet-3}

MetNet-3~\cite{MetNetv3} employs an evaluation protocol that differs fundamentally from ours, in particular by interleaving train and test data. As neither the source code nor forecasts for MetNet-3 are publicly available, a direct, fully controlled comparison is not possible. We therefore made a careful attempt to reproduce their evaluation protocol and compare our results against the figures reported in~\citet{MetNetv3}; see Fig.~\ref{fig:metnet}. Despite these efforts, the HRRR baseline values reported in~\citet{MetNetv3} are substantially higher compared to those obtained in our reproduction attempt. These discrepancies persisted even after discussion with the authors of~\citet{MetNetv3}. As a result, a strict quantitative comparison between MetNet-3 and FitzRoy remains challenging.
Nevertheless, we retrained a FitzRoy model while excluding the full test period defined in~\citet{MetNetv3} from training. The results shown in Fig.~\ref{fig:metnet} indicate consistently higher CSI values for FitzRoy at medium to longer lead times. We expect this performance advantage to increase further under a fully consistent evaluation protocol.
Finally, visual inspection of the forecast examples presented in~\citet{MetNetv3} suggests that MetNet-3 produces comparatively smooth intensity fields. Such smoothing can reduce the impact of the ``double penalty'' effect, potentially inflating skill scores. In contrast, FitzRoy achieves high predictive skill while preserving realistic spatial structures, which is essential for practical forecasting applications. We emphasize that this interpretation is necessarily limited by the lack of publicly available MetNet-3 forecasts and code.

%% file: tex/suppl_figures.tex
\begin{figure}
    \centering
    \includegraphics[width=\textwidth]{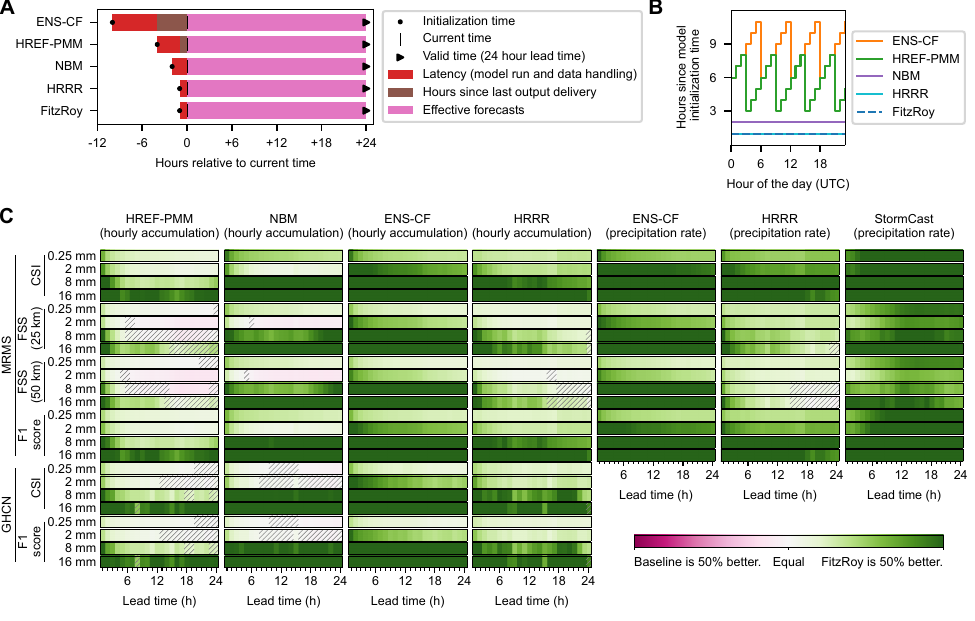}
    \caption{
        \textbf{Latency-aware evaluation of intensity forecasts.} 
        (\textbf{A})~In a real-time operational setting, model latency (red) must be accounted for when producing effective forecasts up to 24 hours ahead (pink). In addition, non-hourly model runs must be taken into account when providing forecasts for every hour of the day~(brown). Earlier model initialization times lead to reduced effective forecast quality at a given valid time. This highlights the practical advantages of low-latency ML models such as FitzRoy in real-time operations, where a lower latency translates into even higher relative skill compared to baseline models.
        We achieve a sub-hour latency of FitzRoy in a real-time setting by using ``MultiSensor\_QPE\_01H\_Pass1'' with a 20 min latency~\cite{mrms}, together with the most recent available NBM and GFS runs as inputs. Requested lead times are adjusted according to the time differences between the current time and the respective model initialization times. 
        The encoding stage of FitzRoy requires approximately 2 seconds, whereas the diffusion process for a given lead time requires approximately 20 seconds on a single A100 GPU. Diffusion processes for different lead times can be executed in parallel.
        (\textbf{B})~Elapsed hours since model initialization time for each hour of the day. For example, at 04:00~UTC, the most recent available HREF-PMM output is 4 hours old; thus, the 5--28 hour forecasts from that model run are deployed as effective real-time operational 1--24 hour forecasts.  
        (\textbf{C})~Scorecards for hourly accumulated precipitation and instantaneous precipitation rate; cf.\ Supplementary Fig.~\ref{fig:nature_1h_tp_oper}--\ref{fig:stormcast_prate_oper}. Hatched cells indicate statistically not significant differences (using a two-sided test with a 5\% significance level~\cite{Hamill1999}).
    }
    \label{fig:operational_setting}
\end{figure}

\begin{figure}
    \centering
    \includegraphics[width=\textwidth]{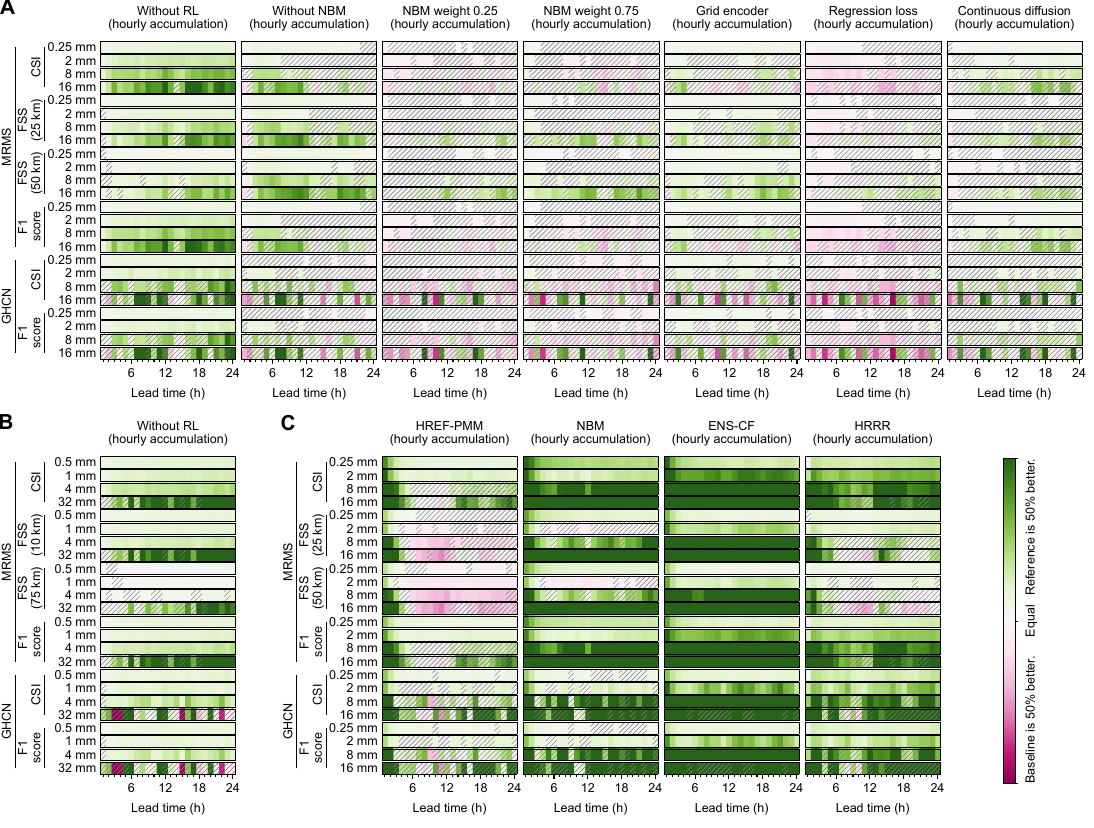}
    \caption{\textbf{Analysis of model components.}
    (\textbf{A})~FitzRoy compared to various model variants. Reinforcement learning provides the largest performance gains. The inclusion of NBM is the second-largest source of improvement, particularly for heavier rain categories. The results are robust to the choice of NBM fusion weight. Using a perceiver-based architecture rather than a grid-based one and discrete rather than continuous diffusion yields marginal but statistically significant gains. Using cross-entropy instead of regression performs similarly but cross-entropy enables probabilistic reasoning beyond regression.
    (\textbf{B})~Robustness to the RL reward definition. We evaluate the model across intensity thresholds and area sizes not used in the RL reward. FitzRoy generalizes strongly, showing even larger gains for several unseen categories, especially for heavier rain.
    (\textbf{C})~FitzRoy without NBM input compared to baseline approaches. The model without NBM performs better than NBM across all metrics except 2-mm CSI and FSS. Furthermore, FitzRoy without NBM outperforms ENS-CF on all metrics, and it surpasses or matches HRRR on all metrics. In all scorecards, hatched cells indicate differences that are not statistically significant (two-sided test, 5\% significance level~\cite{Hamill1999}).} 
     \label{fig:ablations}
\end{figure}

\begin{figure}
    \centering
    \includegraphics[width=\textwidth]{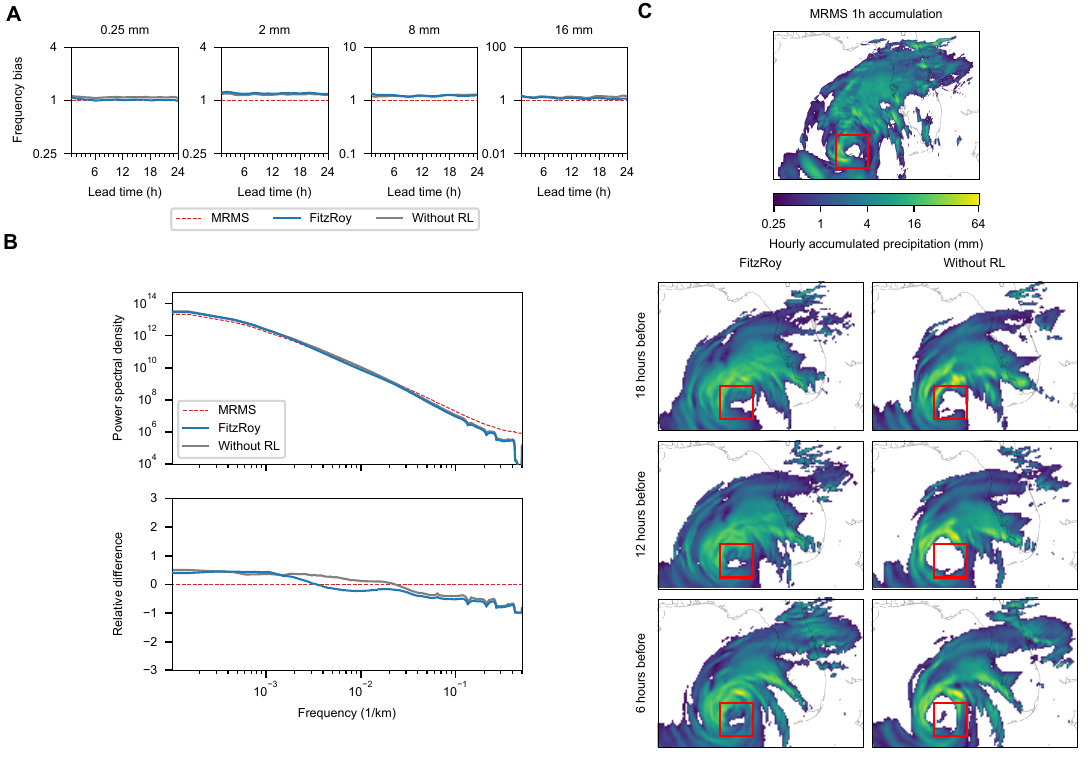}
    \caption{\textbf{Effect of reinforcement learning on forecast realism.}
    (\textbf{A})~Frequency bias as a function of lead time. Both FitzRoy with and without RL exhibit low frequency bias across all lead times.
    (\textbf{B})~Top:~Mean power spectral density of hourly precipitation accumulation as a function of spatial frequency. Bottom:~Relative difference with respect to MRMS, where curves closer to zero indicate better performance. FitzRoy yields similar spectral characteristics with and without RL.
    (\textbf{C})~Hurricane Milton case study (12:00 UTC October 9th, 2024). Both model variants produce spatially realistic precipitation fields, while our final RL-fine-tuned model better captures several structures present in the observations.}
    \label{fig:rl_realism}
\end{figure}

\begin{figure}
    \centering
    \includegraphics[width=\textwidth]{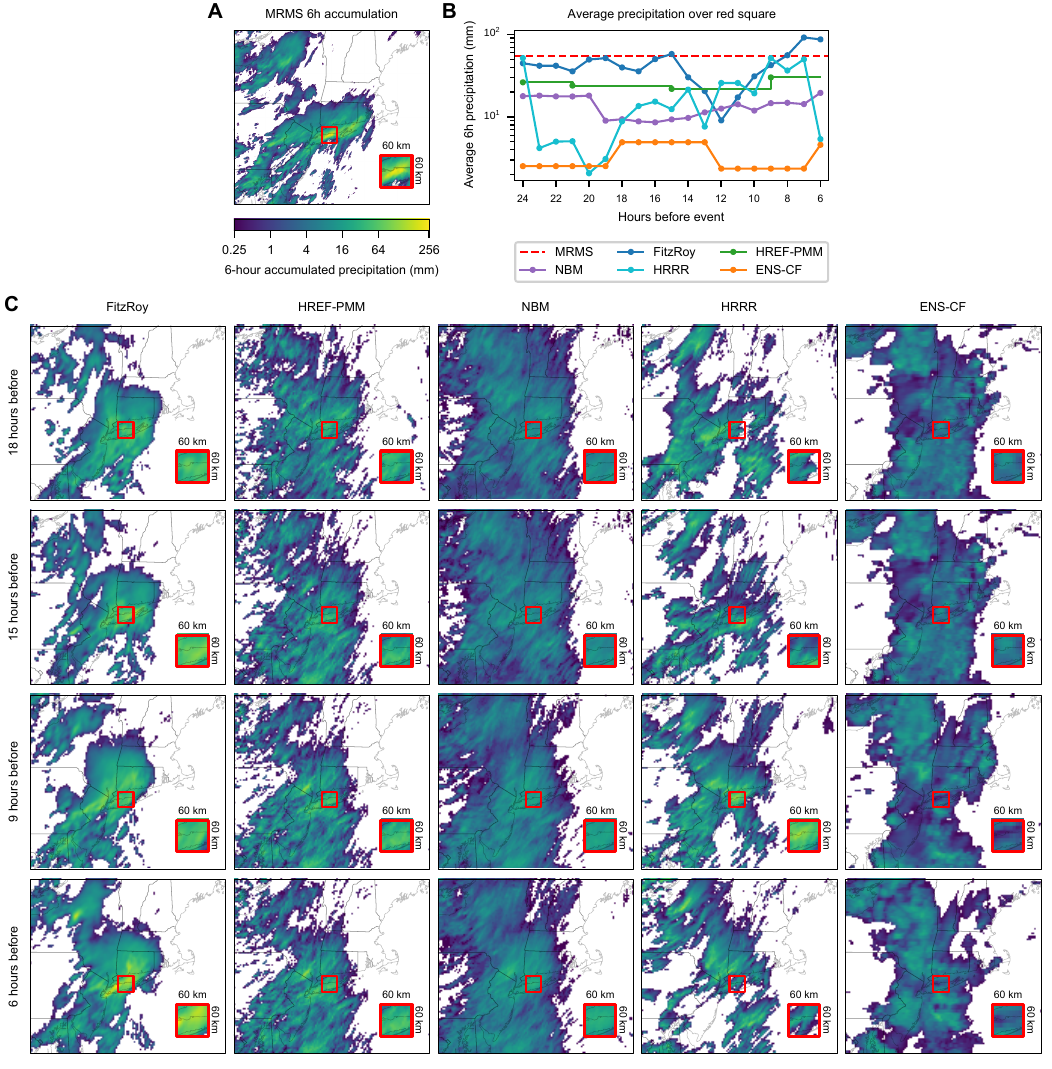}
    \caption{
    \textbf{Long Island, New York flooding.} 
    (\textbf{A})~6-hour observed accumulated precipitation from radar between 00:00 and 06:00 UTC on 19 August 2024 in the northeastern US.  
    (\textbf{B})~Area-averaged observed and forecasted accumulations over the red box shown in panel A (Lat: $40.63^\circ$ -- $41.17^\circ$, Lon: $-73.46^\circ$ -- $ -72.74^\circ$) for different initialization times. 
    (\textbf{C})~6-hour accumulation forecasts for different initialization times (lead times). Although no models had perfect forecasts of these storms, FitzRoy provided earlier indication for the potential of large rainfall accumulations on Long Island, where significant flooding occurred.
    }
    \label{fig:case-study-LI}
    \vspace{1cm}
\end{figure}

\begin{figure}
    \centering
    \includegraphics[width=\textwidth]{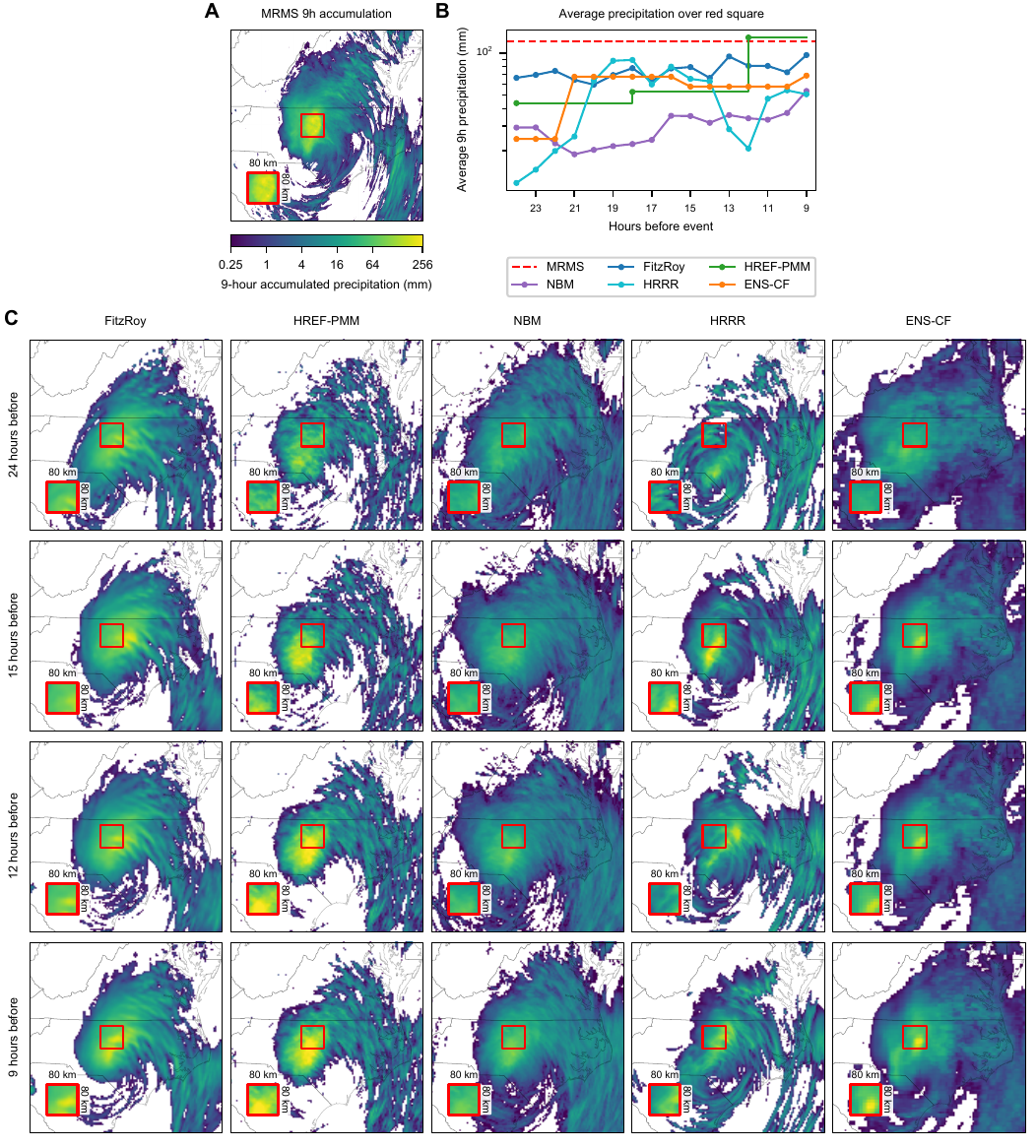}
    \caption{
    \textbf{Remnants of Tropical Storm Chantal.}
    (\textbf{A})~9-hour observed accumulated precipitation from radar between 18:00 and 03:00 UTC on 6--7 July 2025 in North Carolina.  
    (\textbf{B})~Area-averaged observed and forecasted accumulations over the red box shown in panel A (Lat: $35.64^\circ$--$36.36^\circ$, Lon: $-79.74^\circ$--$-78.86^\circ$) for different initialization times. 
    (\textbf{C})~9-hour accumulation forecasts for different initialization times (lead times). FitzRoy consistently identifies the extreme accumulation magnitude across multiple initialization times, whereas operational baselines exhibit significant underprediction and higher run-to-run variability.}
    \label{fig:case-study-chantal}
    \vspace{1cm}
\end{figure}

\begin{figure}
    \centering
    \includegraphics[width=\textwidth]{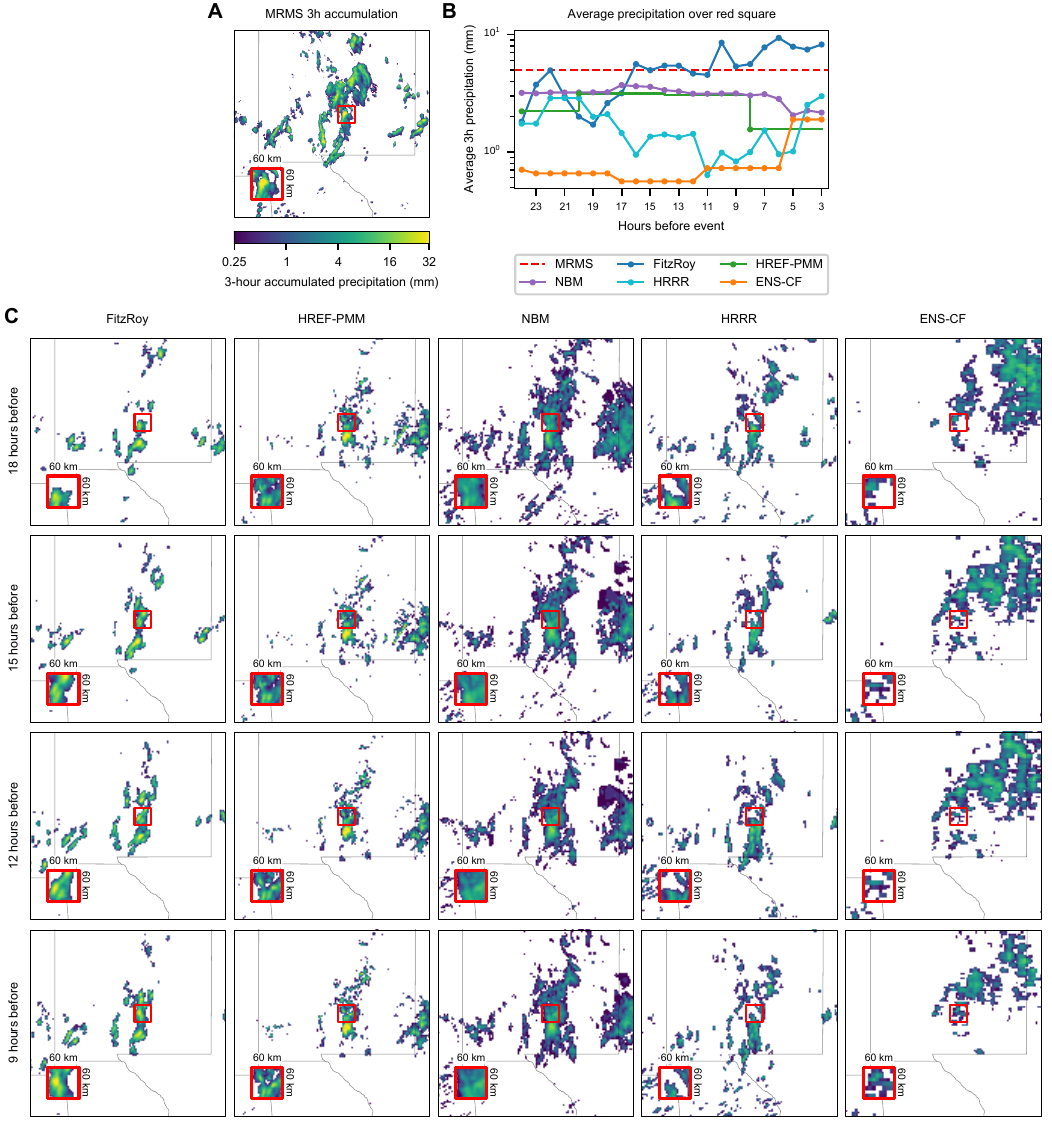}
    \caption{
    \textbf{Ruidoso, New Mexico flooding.} 
    (\textbf{A})~3-hour observed accumulated precipitation from radar between 21:00 and 00:00 UTC on 8 July 2025 in the state of New Mexico.  
    (\textbf{B})~Area-averaged observed and forecasted accumulations over the red box shown in panel A (Lat: $33.03^\circ$ -- $33.57^\circ$, Lon: $-106^\circ$ -- $ -105.35^\circ$) for different initialization times. 
    (\textbf{C})~3-hour accumulation forecasts for different initialization times (lead times). FitzRoy's forecasts of this very localized cluster of rainstorms were a closer match to the observed precipitation.}
    \label{fig:case-study-ruidoso}
    \vspace{1cm}
\end{figure}


\begin{figure}
    \centering
    \includegraphics[width=11.4cm]{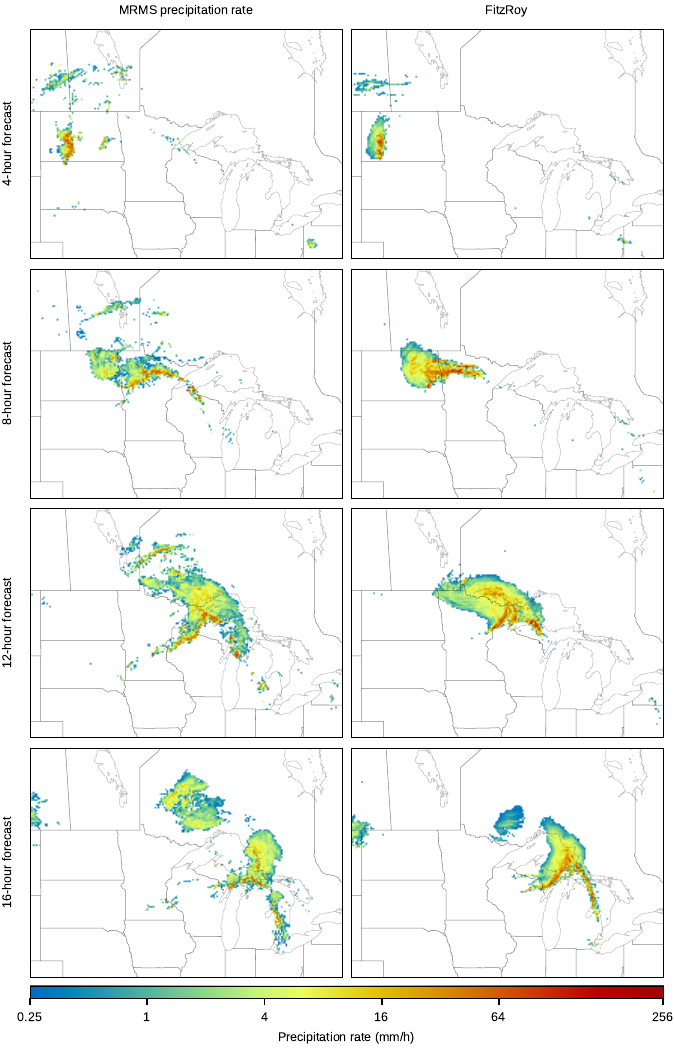}
    \caption{\textbf{Northern US Derecho.} MRMS and forecasted precipitation rate (Lat: $40^\circ$ -- $55^\circ$, Lon: $-105^\circ$ -- $ -77^\circ$) for different forecast lead times initialized at 22:00 UTC June 20th, 2025 (latency-adjusted). FitzRoy successfully forecasts the squall line and structure of future observations.}
    \label{fig:case-study-derecho}
\end{figure}

\begin{figure}
    \centering
    \includegraphics[width=11.4cm]{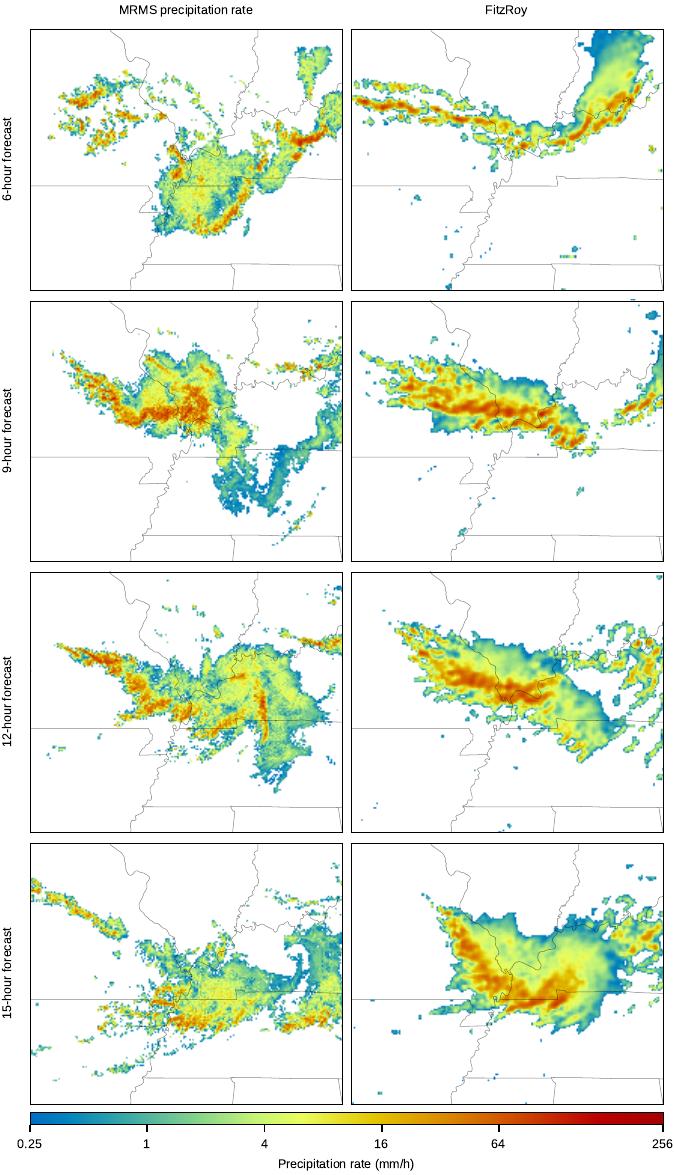}
    \caption{\textbf{Missouri flood.} MRMS and forecasted precipitation rate (Lat: $34.5^\circ$ -- $39.5^\circ$, Lon: $-93^\circ$ -- $-85.5^\circ$) for different forecast lead times initialized at 21:00 UTC July 9th, 2026 (latency-adjusted).}
    \label{fig:case-study-missouri-prate}
\end{figure}

\begin{figure}
    \centering
    \includegraphics[width=\textwidth]{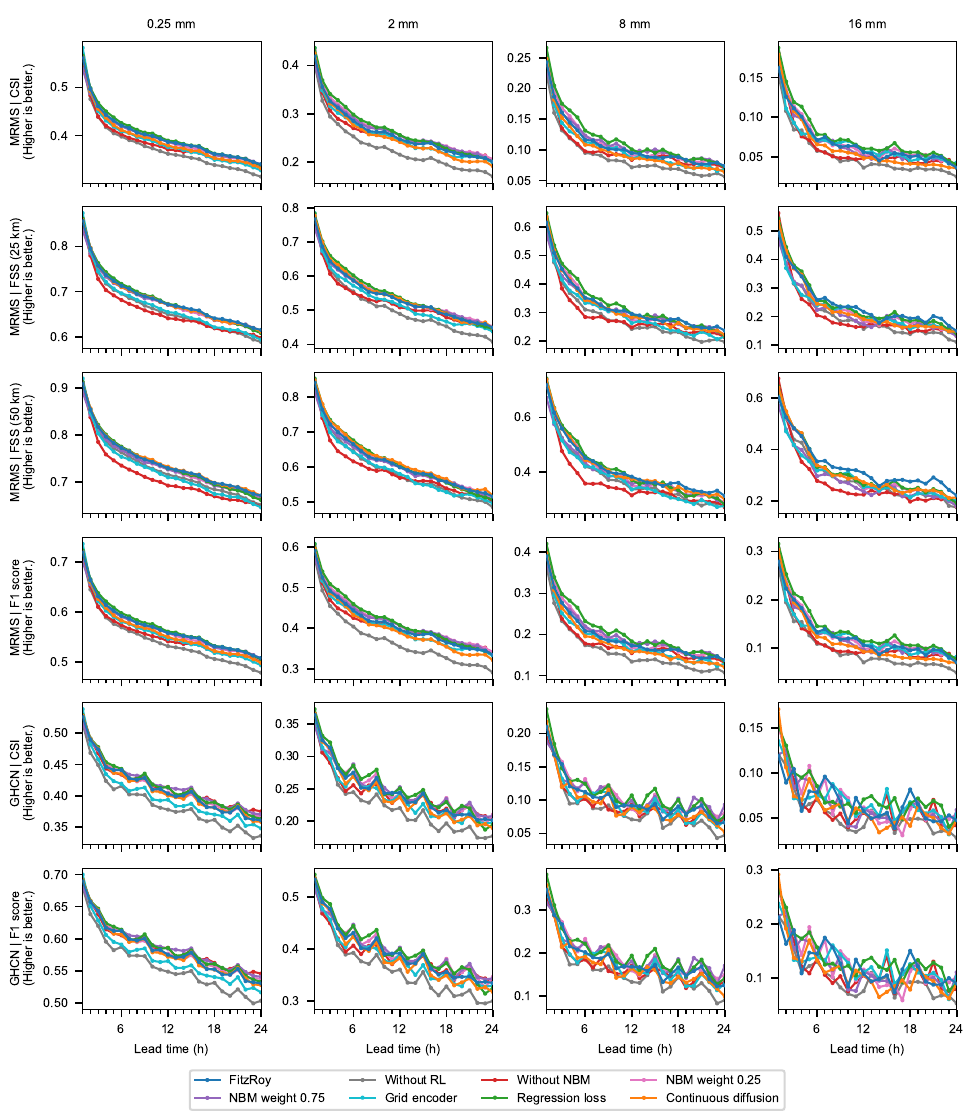}
    \caption{\textbf{Evaluation of model components.} For each model variant, we evaluate four runs per day for the given evaluation time range: 00, 06, 12, and 18 UTC.}
\end{figure}

\begin{figure}
    \centering
    \includegraphics[width=\textwidth]{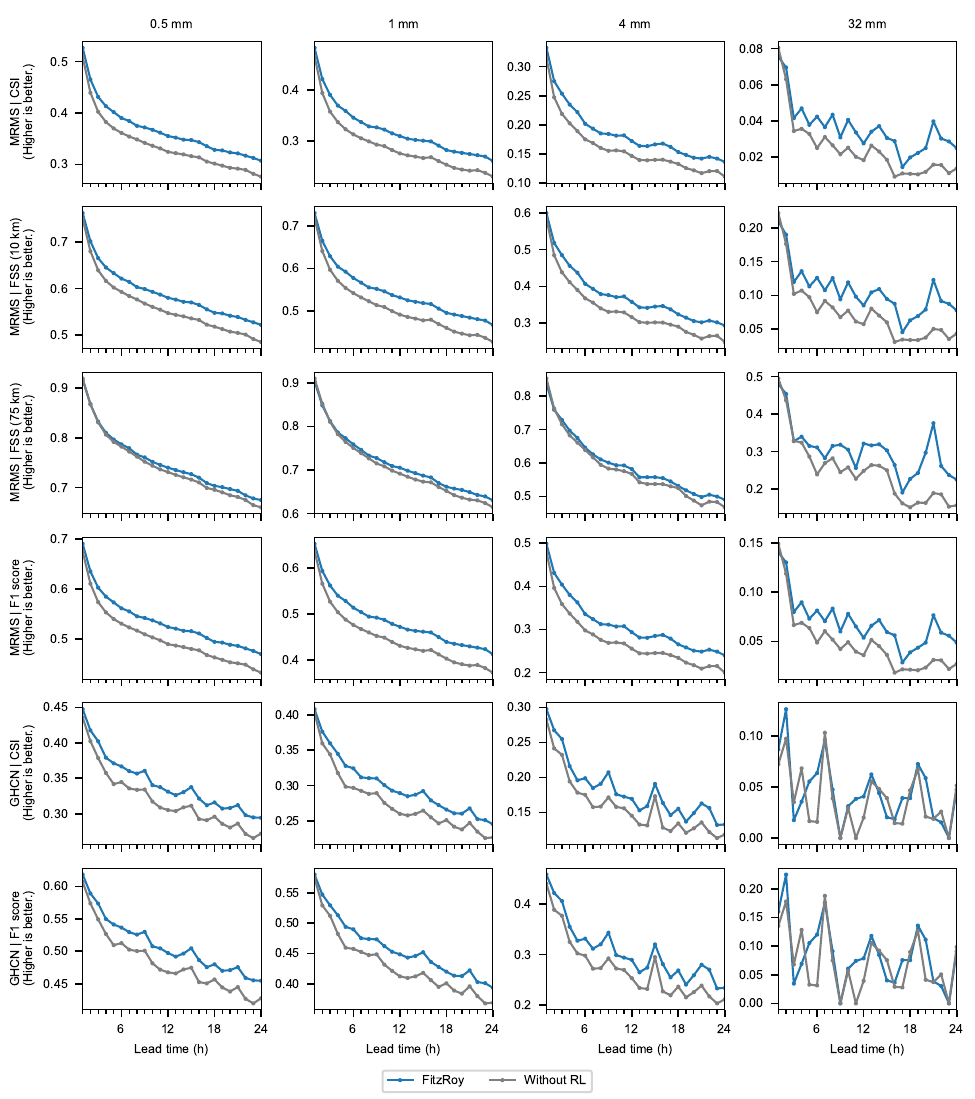}
    \caption{\textbf{Additional evaluation of the effect of reinforcement learning.} For each model variant, we evaluate four runs per day for the given evaluation time range: 00, 06, 12, and 18 UTC.}
\end{figure}

\begin{figure}
    \centering
    \includegraphics[width=\textwidth]{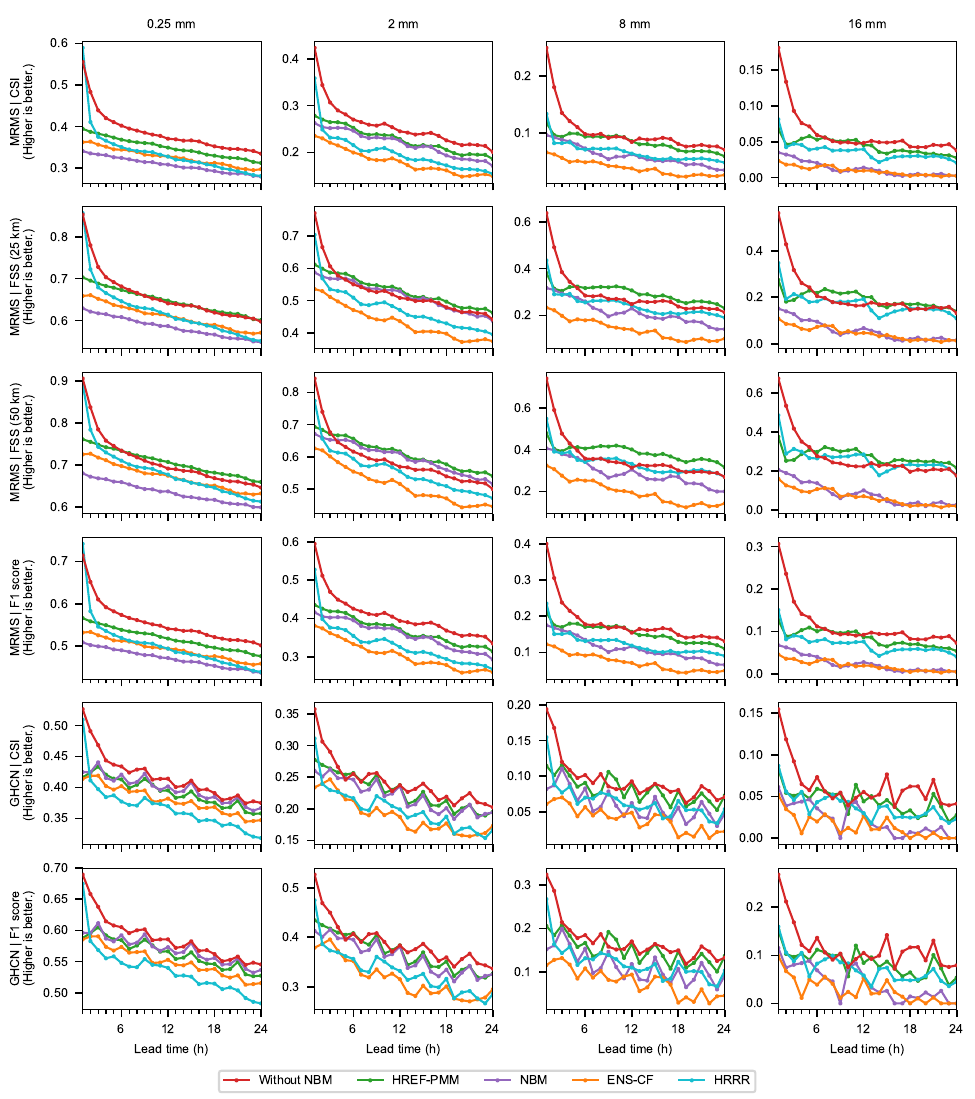}
    \caption{\textbf{Evaluation of a model variant without NBM.} For each model, we evaluate four runs per day for the given evaluation time range: 00, 06, 12, and 18 UTC.}
\end{figure}

\begin{figure}
    \centering
    \includegraphics[width=\textwidth]{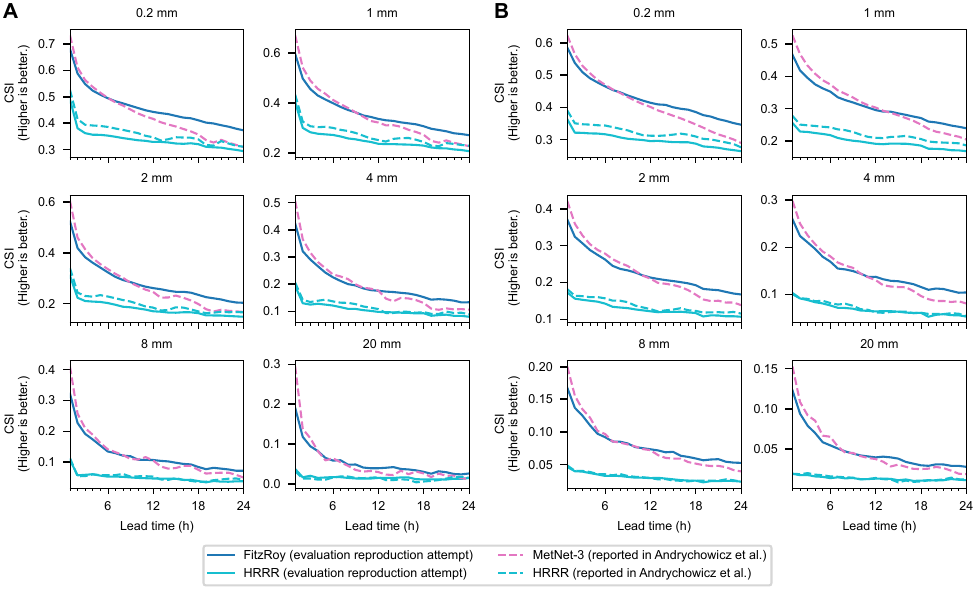}
    \caption{
        \textbf{Attempted comparison with MetNet-3.}
        Since neither source code nor operational forecasts from MetNet-3~\cite{MetNetv3} are publicly available, we follow the evaluation protocol described in~\citet{MetNetv3} (solid lines) and compare against the values reported in~\citet{MetNetv3}~(dashed lines).
        Reported values for HRRR~(dashed, cyan) are substantially higher than those obtained in our reproduction attempt~(solid, cyan), which complicates a direct comparison between MetNet-3~(dashed, pink) and FitzRoy~(solid, blue). Despite those discrepancies, which persisted even after discussion with the authors of~\citet{MetNetv3}, the FitzRoy curves (solid, blue) show higher CSI values at medium to longer lead times. We expect this performance advantage to increase further under a fully consistent evaluation procedure.
        (\textbf{A})~Hourly accumulated precipitation.
        (\textbf{B})~Instantaneous precipitation rate.
    }
    \label{fig:metnet}
\end{figure}

\begin{figure}
    \centering
    \includegraphics[width=\textwidth]{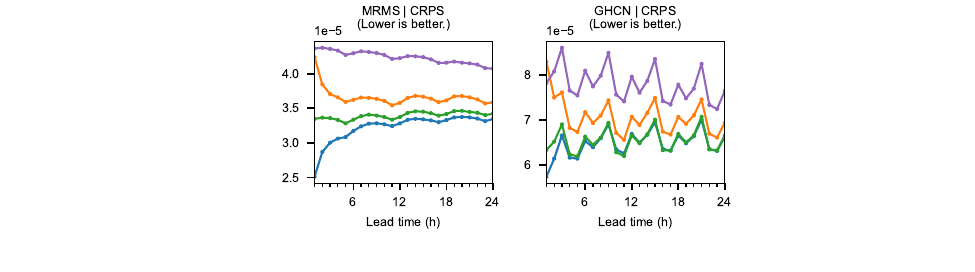}\\[-14mm]
    \includegraphics[width=\textwidth]{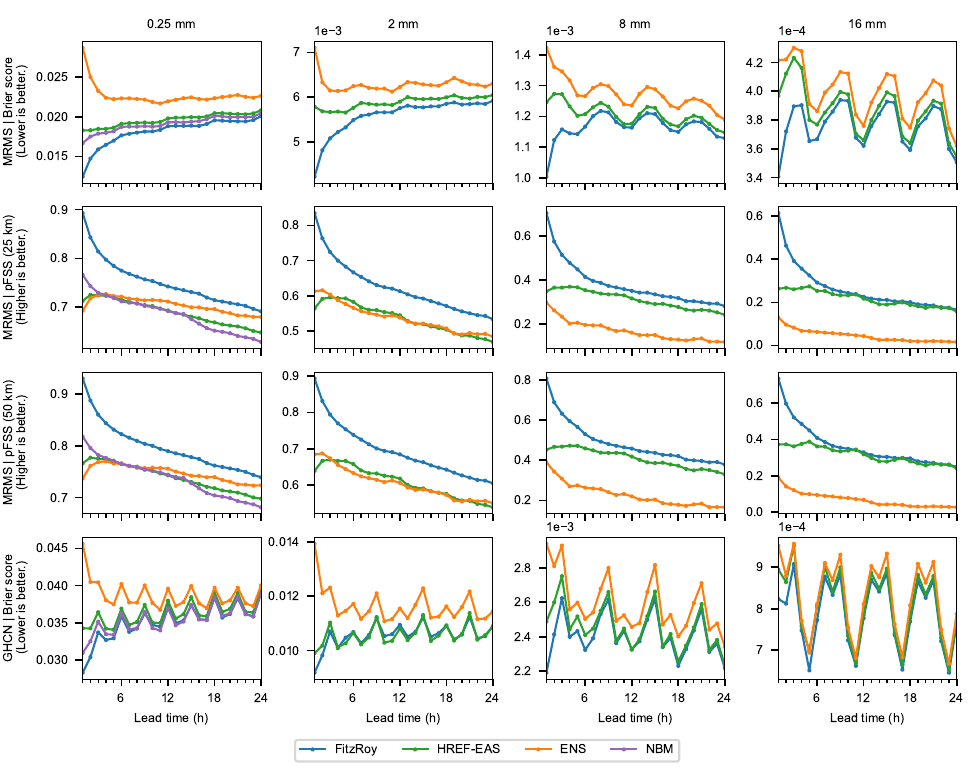}
    \caption{\textbf{Evaluation of hourly probability of rain.} For each model, we evaluate four runs per day for the given evaluation time range: 00, 06, 12, and 18 UTC.}
\end{figure}

\begin{figure}
    \centering
    \includegraphics[width=\textwidth]{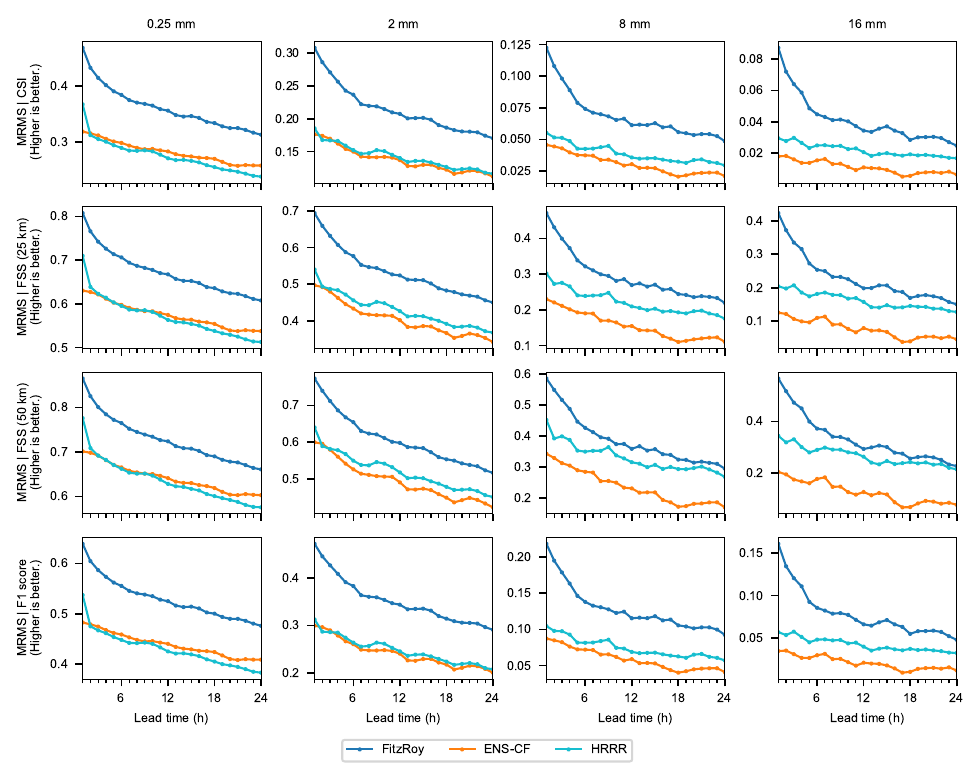}
    \caption{\textbf{Evaluation of instantaneous precipitation rate.} For each model, we evaluate four runs per day for the given evaluation time range: 00, 06, 12, and 18 UTC.}
\end{figure}

\begin{figure}
    \centering
    \includegraphics[width=\textwidth]{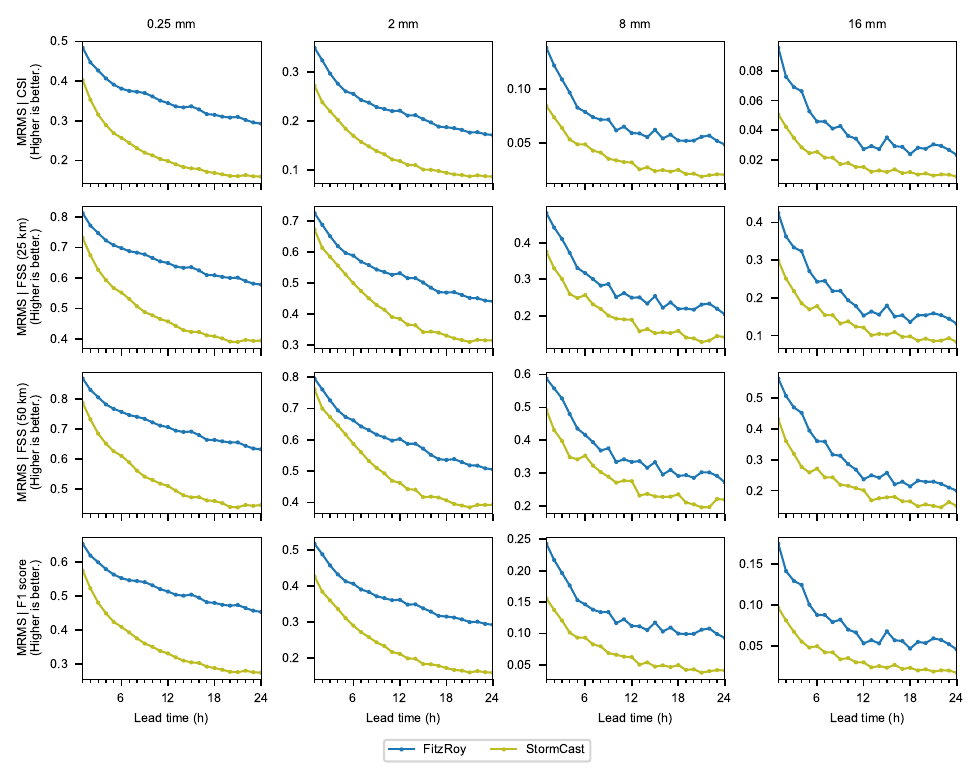}
    \caption{\textbf{Evaluation of instantaneous precipitation rate w.r.t.\ StormCast~\cite{StormCast}.} For each model, we evaluate four runs per day for the given evaluation time range: 00, 06, 12, and 18 UTC. Since StormCast only operates on a 1,536 km $\times$ 1,920 km CONUS subregion~\cite{StormCast}, we use the same subregion here.}
\end{figure}

\begin{figure}
    \centering
    \includegraphics[width=\textwidth]{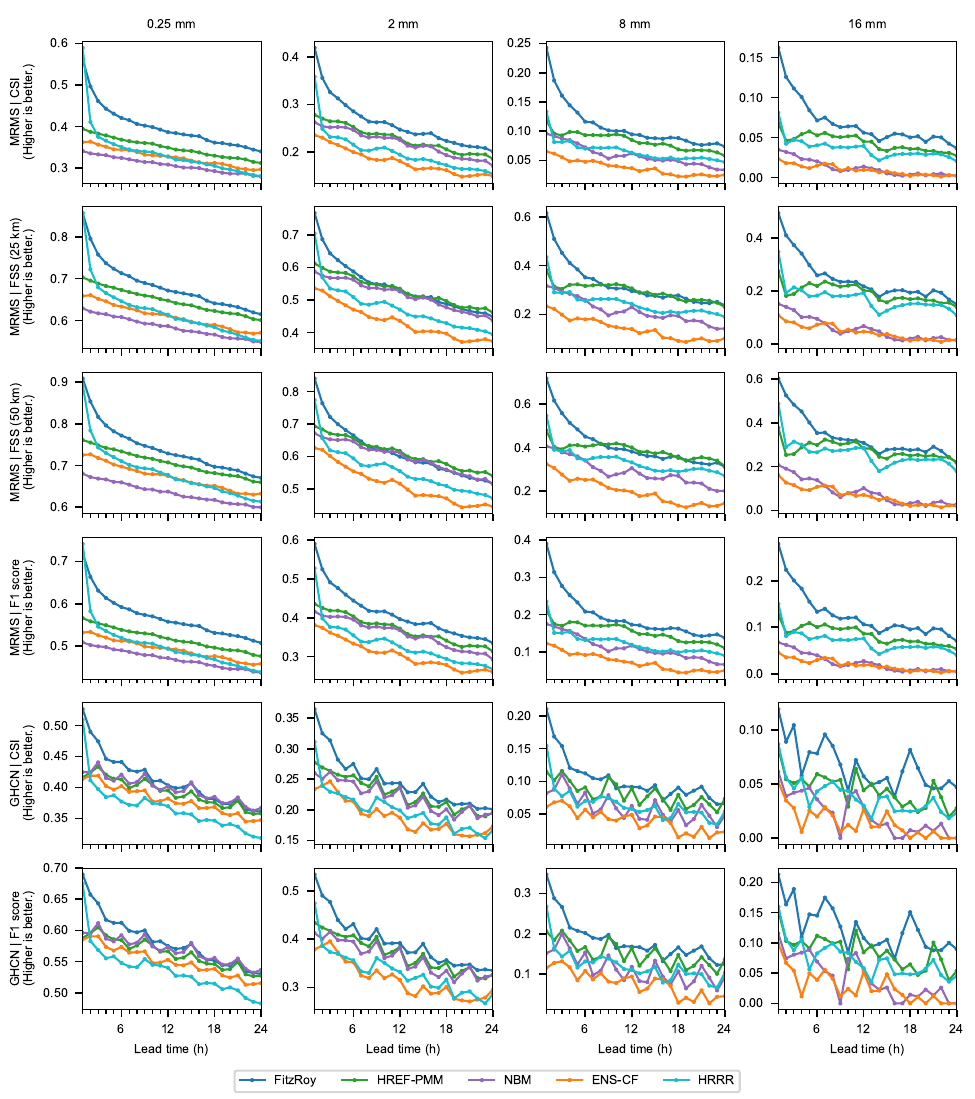}
    \caption{\textbf{Evaluation of hourly precipitation accumulation.} For each model, we evaluate four runs per day for the given evaluation time range: 00, 06, 12, and 18 UTC.}
\end{figure}

\begin{figure}
    \centering
    \includegraphics[width=\textwidth]{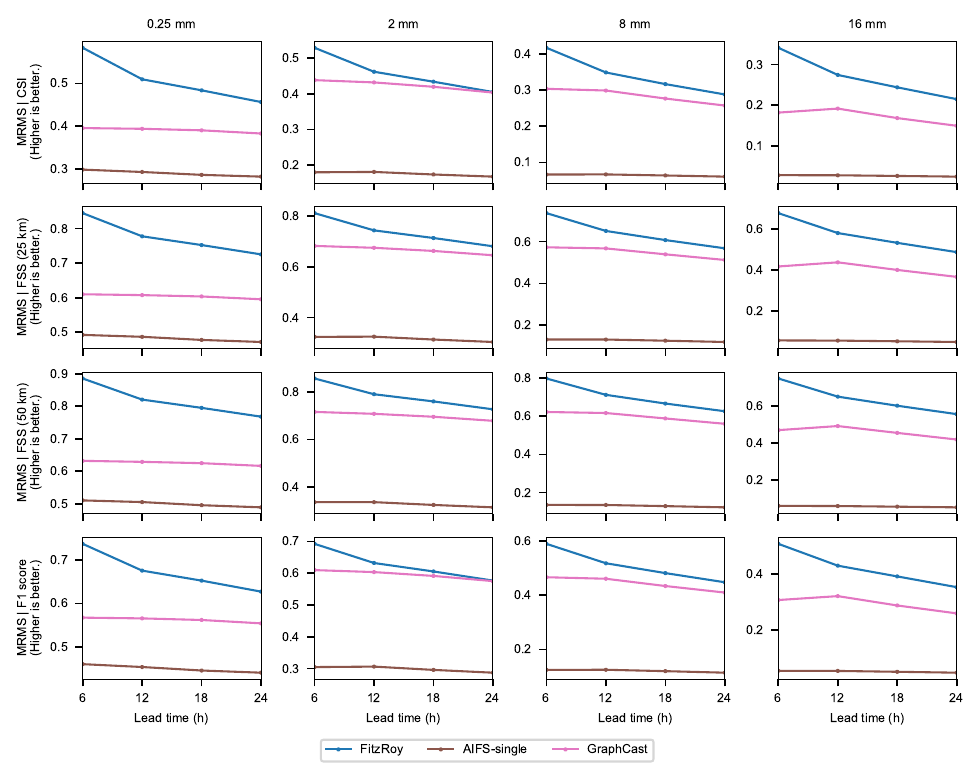}
    \caption{\textbf{Evaluation of 6-hour precipitation accumulation compared to existing ML prediction models.} For each model, we evaluate four runs per day for the given evaluation time range: 00, 06, 12, and 18 UTC, against current state-of-the-art ML weather prediction models AIFS-single and GraphCast. FitzRoy outperforms these models across all metrics and intensities analyzed.}
\end{figure}

\begin{figure}
    \centering
    \includegraphics[width=\textwidth]{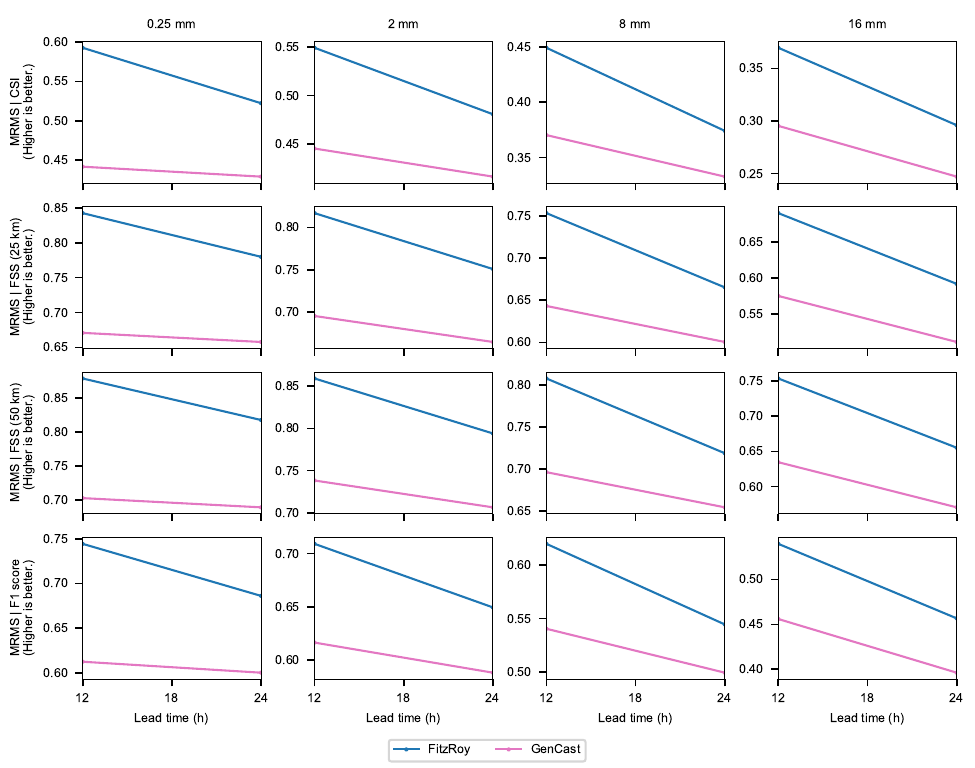}
    \caption{\textbf{Evaluation of 12-hour precipitation accumulation compared to GenCast.} For each model, we evaluate two runs per day for the given evaluation time range: 00 and 12 UTC against the GenCast model. FitzRoy outperforms GenCast across all metrics and intensities analyzed.}
\end{figure}

\begin{figure}
    \centering
    \includegraphics[width=\textwidth]{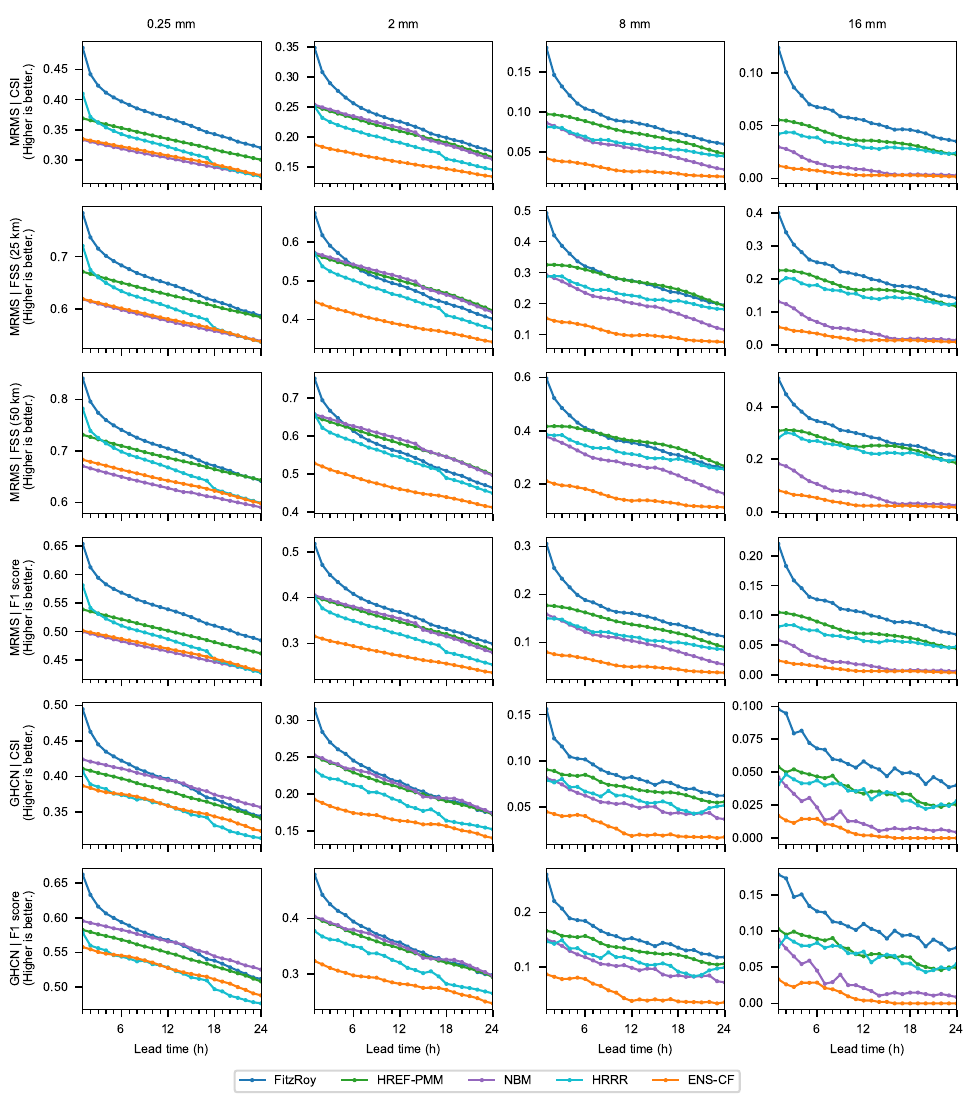}
    \caption{\textbf{Latency-aware evaluation of hourly precipitation accumulation.} For each model, we evaluate on every hour for the given evaluation time range where we take into account the operational latencies of the models (see Fig.~\ref{fig:operational_setting}).}
    \label{fig:nature_1h_tp_oper}
\end{figure}

\begin{figure}
    \centering
    \includegraphics[width=\textwidth]{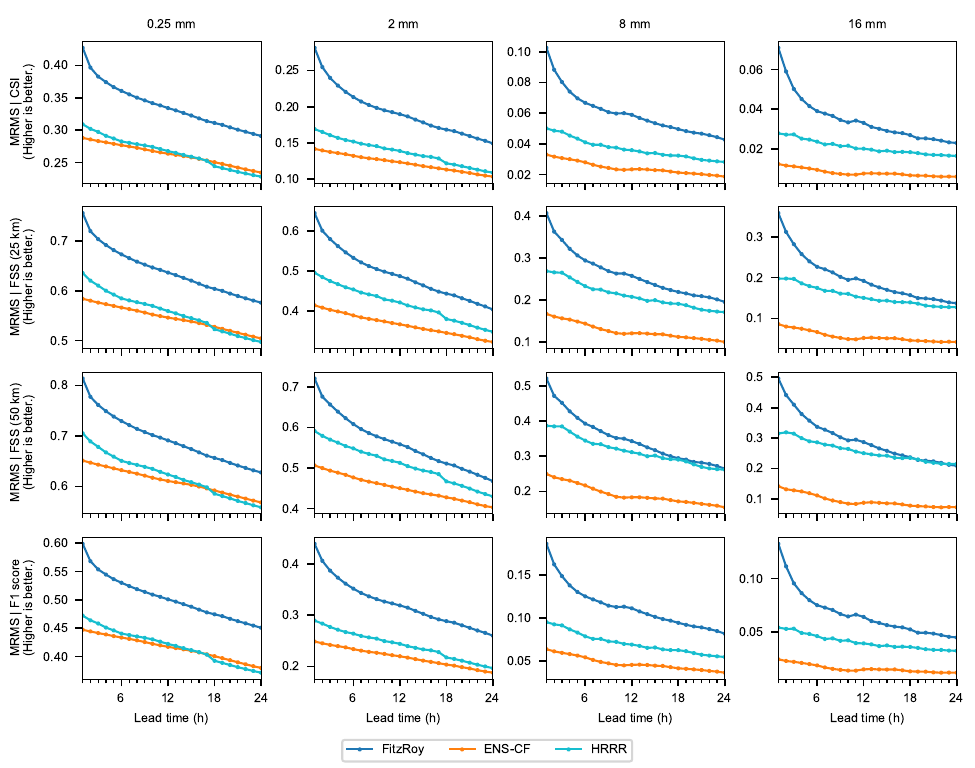}
    \caption{\textbf{Latency-aware evaluation of instantaneous precipitation rate.} For each model, we evaluate on every hour for the given evaluation time range where we take into account the operational latencies of the models (see Fig.~\ref{fig:operational_setting}).}
    \label{fig:nature_1h_prate_oper}
\end{figure}

\begin{figure}
    \centering
    \includegraphics[width=\textwidth]{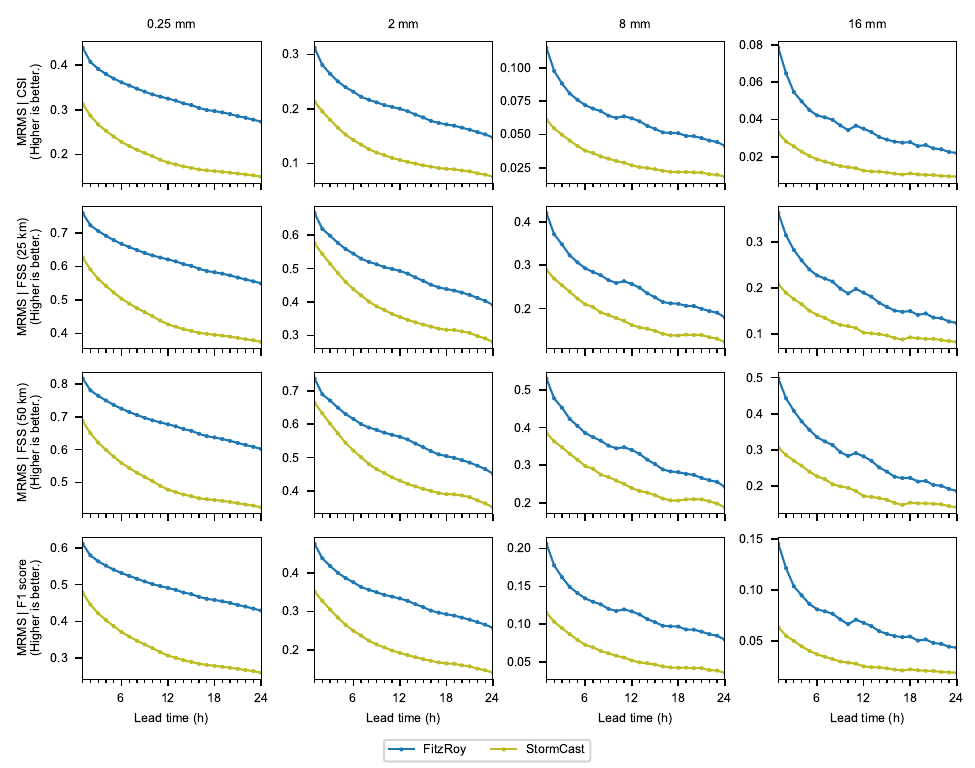}
    \caption{\textbf{Latency-aware evaluation of instantaneous precipitation rate w.r.t.\ StormCast.} For each model, we evaluate on every hour for the given evaluation time range where we take into account the operational latencies of the models. Since StormCast only operates on a 1,536 km $\times$ 1,920 km CONUS subregion~\cite{StormCast}, we evaluate here on the same subregion.}
    \label{fig:stormcast_prate_oper}
\end{figure}

%% file: tex/suppl_tables.tex
\begin{table}
\centering
\caption{\textbf{Computational cost of the FitzRoy training pipeline.} We report the number of unique hourly records used by each stage, the corresponding number of equivalent passes through those records, the hardware configuration, wall-clock training time, and total GPU-hours. Equivalent passes are used instead of conventional epochs because initialization times are sampled with replacement and paired with dynamically sampled forecast lead times.}
\begin{tabular}{lrrrrr}
\toprule
Training stage & Hourly records & Equivalent passes & Hardware & Wall time & GPU-hours \\
\midrule
Encoder + probability decoder & 302K & 20   & 64 A100 & 2 days & 3,072 \\
Tokenizer                     & 70K  & 71   & 32 A100 & 1 day  & 768 \\
MaskGIT pretraining           & 167K & 48   & 64 A100 & 6 days & 9,216 \\
MaskGIT RL fine-tuning        & 39K  & 0.35 & 48 B200 & 1 day  & 1,152 \\
\bottomrule
\end{tabular}
\label{tab:training_cost}
\end{table}

\begin{table}
    \centering
    \caption{Output characteristics of FitzRoy and baseline models for precipitation forecasts up to 24 hours. StormCast operates on a 1,536 km $\times$ 1,920 km CONUS subregion~\cite{StormCast}. (NWP: Numerical Weather Prediction, ML: Machine Learning, CONUS: Continental United States.)}
    \label{tab:baselines}
    \vspace{2mm}
    \footnotesize
    \renewcommand{\arraystretch}{1.2}
    \setlength{\tabcolsep}{5pt}
    \begin{tabular}{@{}lccccccc@{}}
        \hline
        \textbf{Model} & \textbf{Type} & \textbf{Region} & 
        \makecell{\textbf{Spatial}\\\textbf{Resolution}} & 
        \makecell{\textbf{Forecast}\\\textbf{Frequency}} & 
        \makecell{\textbf{Precipitation}\\\textbf{Accumulation}} &
        \makecell{\textbf{Precipitation}\\\textbf{Rate}} &
        \makecell{\textbf{Probability of}\\\textbf{Precipitation}} \\
        \hline
        HREF & NWP & CONUS & 3 km & 6 hours & hourly & --- & hourly \\ 
        NBM & NWP & CONUS & 2.5 km & 1 hour & hourly & --- & hourly (rain/no-rain)\\ 
        ENS & NWP & global & 0.25\textdegree & 6 hours & hourly & hourly & hourly \\ 
        HRRR & NWP & CONUS & 3 km & 1 hour & hourly & hourly & --- \\ 
        \hline
        AIFS-single & ML & global & 0.25\textdegree & 6 hours & 6 hours & --- & --- \\ 
        GraphCast & ML & global & 0.25\textdegree & 6 hours & 6 hours & --- & --- \\ 
        GenCast & ML & global & 0.25\textdegree & 12 hours & 12 hours & --- & 12 hours \\ 
        StormCast & ML & subregion & 3 km & 1 hour & --- & hourly & --- \\ 
        \hline
        FitzRoy & ML & CONUS & 1 km & 1 hour & hourly & hourly & hourly \\ 
        \hline
    \end{tabular}
    \renewcommand{\arraystretch}{1}
    \setlength{\tabcolsep}{6pt}
\end{table}

\begin{table}
    \centering
    \caption{Training inputs for the encoder and pointwise probability decoder.}
    \label{tab:train_inputs}
    \vspace{2mm}
    \footnotesize
    \renewcommand{\arraystretch}{1.2}
    \setlength{\tabcolsep}{5pt}
    \begin{tabular}{@{}lc@{}}
        \hline
        \textbf{Input Source} & \textbf{\#Channels} \\
        \hline
        ERA5 atmospheric variables & 21 \\
        MRMS hourly accumulation & 1 \\
        MRMS precipitation type & 1 \\
        NBM hourly accumulation forecast & 1 \\
        NBM precipitation type forecast & 1 \\
        \hline
    \end{tabular}
    \renewcommand{\arraystretch}{1}
    \setlength{\tabcolsep}{6pt}
\end{table}

\begin{table}
    \centering
    \caption{Training targets and loss functions for the encoder and probability decoder.}
    \label{tab:train_targets}
    \vspace{2mm}
    \footnotesize
    \renewcommand{\arraystretch}{1.2}
    \setlength{\tabcolsep}{5pt}
    \begin{tabular}{@{}lc@{}}
        \hline
        \textbf{Target Variable} & \textbf{Loss Function} \\
        \hline
        ERA5 atmospheric variables & Mean squared error \\
        ERA5 hourly accumulation & Cross entropy (401 bins) \\
        MRMS hourly accumulation & Cross entropy (401 bins) \\
        MRMS precipitation type & Cross entropy (4 classes) \\
        \hline
    \end{tabular}
    \renewcommand{\arraystretch}{1}
    \setlength{\tabcolsep}{6pt}
\end{table}

\begin{table}
    \centering
    \caption{ERA5 variables at 0.25\textdegree\ resolution, cropped to CONUS with a 7.5\textdegree\ padding.}
    \label{tab:era5-channels}
    \vspace{2mm}
    \footnotesize
    \renewcommand{\arraystretch}{1.2}
    \setlength{\tabcolsep}{6pt}
    \begin{tabular}{@{}ll@{}}
        \hline
        \textbf{Variable} & \textbf{Level} \\
        \hline
        geopotential                    & 1000 hPa \\
        u\_component\_of\_wind          & 1000 hPa \\
        v\_component\_of\_wind          & 1000 hPa \\
        geopotential                    & 850 hPa \\
        relative\_humidity              & 850 hPa \\
        temperature                     & 850 hPa \\
        u\_component\_of\_wind          & 850 hPa \\
        v\_component\_of\_wind          & 850 hPa \\
        geopotential                    & 500 hPa \\
        relative\_humidity              & 500 hPa \\
        temperature                     & 500 hPa \\
        u\_component\_of\_wind          & 500 hPa \\
        v\_component\_of\_wind          & 500 hPa \\
        geopotential                    & 50 hPa \\
        mean\_sea\_level\_pressure      & single-level \\
        surface\_pressure               & single-level \\
        2m\_temperature                 & single-level \\
        total\_column\_water\_vapour    & single-level \\
        total\_precipitation            & single-level \\
        10m\_u\_component\_of\_wind     & single-level \\
        10m\_v\_component\_of\_wind     & single-level \\
        \hline
    \end{tabular}
    \renewcommand{\arraystretch}{1}
\end{table}

\begin{table}
    \centering
    \caption{Precipitation data sources and variables used by FitzRoy. MRMS and NBM are gridded data. GHCN provides point observations.}
    \label{tab:non-era5-channels}
    \vspace{2mm}
    \footnotesize
    \renewcommand{\arraystretch}{1.2}
    \setlength{\tabcolsep}{6pt}
    \begin{tabular}{@{}lll@{}}
        \hline
        \textbf{Data Source} & \textbf{Variable} & \textbf{Description} \\
        \hline
        MRMS & MultiSensor\_QPE\_01H\_Pass1 & Multi-sensor accumulation 1-hour (20-minute latency) \\
            & MultiSensor\_QPE\_01H\_Pass2 & Multi-sensor accumulation 1-hour (1-hour latency) \\
            & PrecipRate & Radar precipitation rate \\
            & PrecipFlag & Surface Precipitation Type \\
        \hline
        NBM  & QPF01 & Quantitative precipitation forecast \\
             & PType & Probability of Snow, Rain, Sleet, or Freezing Rain \\
        \hline
        GHCN & precipitation & Total liquid precipitation \\
        \hline
    \end{tabular}
    \renewcommand{\arraystretch}{1}
\end{table}

\begin{table}[ht]
    \centering
    \caption{Hyperparameters for training FitzRoy.}
    \label{tab:hyperparams}
    \vspace{2mm}
    \footnotesize
    \renewcommand{\arraystretch}{1.2}
    \begin{tabular}{@{}lll@{}}
        \hline
        \multirow{12}{*}{\makecell[l]{Encoder and \\ probability decoder}} & Loss weight (ERA5 channels) & 0.1 \\
         & Loss weight (ERA5 hourly accumulation) & 0.1 \\
         & Loss weight (MRMS hourly accumulation) & 1.0 \\
         & Loss weight (MRMS precipitation type) & 0.1 \\
         & Latent dimension ($D$) & 1024 \\
         & Number of query tokens ($N$) & 1024 \\
         & MRMS quality clamp & 0.01 \\
         & Transformer depth ($L$) & 8 \\
         & Number of attention heads & 8 \\
         & Learning rate & 3e-4 \\
         & Batch size & 64 \\
         & EMA weight & 0.999 \\
        \hline
        \multirow{8}{*}{Tokenizer} & Perceptual loss weight & 0.8 \\
         & Quantization dropout & 0.25 \\
         & Patch size & 8 \\
         & Encoder base channels & 128 \\
         & Decoder base channels & 64 \\
         & Learning rate & 1e-4 \\
         & Batch size & 32 \\
         & EMA weight & 0.999 \\
        \hline
        \multirow{18}{*}{Intensity decoder} & Transformer depth & 24 \\
         & Hidden dimension size & 1024 \\
         & Number of attention heads & 16 \\
         & MLP dimension & 4096 \\
         & CFG weight & 3.0 \\
         & Number of decoding steps per trajectory & 32 \\
         & Number of trajectories per RL update & 10 \\
         & RL rain thresholds (mm) & 0.25, 2, 8, 16 \\
         & Hourly accumulation reward weights (FSS, FB) & 0.9, 0.1 \\
         & Hourly accumulation threshold weights & 3, 1, 1, 0 \\
         & Precipitation rate reward weights (FSS, FB) & 0.8, 0.2 \\
         & Precipitation rate threshold weights & 3, 1, 1, 1 \\
         & Policy-gradient loss weight & 10 \\
         & KL-divergence loss weight & 0.1 \\
         & Cross-entropy loss weight & $10^{-4}$ \\
         & Learning rate & 1e-4 \\
         & Batch size & 64 \\
         & EMA weight & 0.999 \\
        \hline
    \end{tabular}
\end{table}